\documentclass{article}

\usepackage{microtype}
\usepackage{graphicx}
\usepackage{subcaption}
\usepackage{booktabs} 

\usepackage{hyperref}

\usepackage[accepted]{icml2026}

\usepackage{amsmath}
\usepackage{amssymb}
\usepackage{mathtools}
\usepackage{amsthm}

\usepackage[most]{tcolorbox}
\newtcolorbox{promptbox}{
  enhanced,
  breakable,
  colback=gray!3,
  colframe=gray!50,
  boxrule=0.6pt,
  arc=2pt,
  left=6pt,
  right=6pt,
  top=6pt,
  bottom=6pt,
  fonttitle=\bfseries,
  title=LLM-as-a-Judge Prompt,
}

\usepackage[capitalize,noabbrev]{cleveref}

\theoremstyle{plain}
\newtheorem{theorem}{Theorem}[section]

\theoremstyle{definition}
\newtheorem{definition}[theorem]{Definition}

\theoremstyle{remark}

\usepackage[disable,textsize=tiny]{todonotes}

\icmltitlerunning{Escaping Mode Collapse in LLM Generation via Geometric Regulation}

\begin{document}


\twocolumn[
    \icmltitle{Escaping Mode Collapse in LLM Generation via Geometric Regulation}



	\icmlsetsymbol{equal}{*}

	\begin{icmlauthorlist}
		\icmlauthor{Xin Du}{xxx,zzz}
		\icmlauthor{Kumiko Tanaka-Ishii}{yyy}
	\end{icmlauthorlist}

    \vbox to 0pt{%
        \vskip 0.04in
        \hbox to \textwidth{\hfil\small\textbf{Code:}
        \href{https://github.com/kduxin/llm-reinforced-mode-regulation}
        {kduxin/llm-reinforced-mode-regulation}\hfil}%
        \vss
    }

	\icmlaffiliation{xxx}{Department of Communications and Computer Engineering, Waseda University, Tokyo, Japan}
	\icmlaffiliation{yyy}{Department of Computer Science and Engineering, Waseda University, Tokyo, Japan}
	\icmlaffiliation{zzz}{Shanghai Research Institute for Intelligent Autonomous Systems, Tongji University, Shanghai, China}

	\icmlcorrespondingauthor{Kumiko Tanaka-Ishii}{kumiko@waseda.jp}

\icmlkeywords{Large Language Models, Neural Decoding, Dynamical Systems, Fractal Dimension, Correlation Dimension, Complexity, Text Generation, Text Quality Evaluation, Mode Collapse, Phase Transition, Complex Systems, Language as a Dynamical System}

	\vskip 0.3in
]



\printAffiliationsAndNotice{}

\begin{abstract}

Mode collapse is a persistent challenge in generative modeling and manifests in
autoregressive text generation as behaviors ranging from explicit looping to
gradual loss of diversity and premature trajectory convergence. We take a
dynamical-systems view and reinterpret mode collapse as reduced state-space
accessibility caused by \emph{geometric collapse}: during generation, the
model's internal trajectory becomes confined to a low-dimensional region of its
representation space. This implies mode collapse is not purely a token-level
phenomenon and cannot be reliably mitigated by symbolic constraints or
probability-only decoding heuristics. Guided by this perspective, we propose
\emph{Reinforced Mode Regulation} (RMR), a lightweight, online state-space
intervention that regulates dominant self-reinforcing directions in the
Transformer value cache (implemented as low-rank damping). Across multiple large
language models, RMR substantially reduces mode collapse and enables stable,
high-quality generation at extremely low entropy rates (down to 0.8 nats/step),
whereas standard decoding typically collapses near 2.0 nats/step.

\end{abstract}

\section{Introduction}

Large language models (LLMs) can generate remarkably fluent text, yet
long-horizon decoding remains fragile: small local biases can accumulate into
macroscopic failure modes. A generation process may gradually lose diversity
and become repetitive, bland, or abruptly trapped in self-reinforcing patterns.
We refer to this family of long-range failures as \emph{mode collapse}. A key
challenge is that mode collapse is hard to define locally or predict from
individual next-token probabilities, even though its consequences dominate
downstream usability in long-form generation.

Much of the decoding literature addresses these failures by modifying the
next-token distribution---through truncation, penalties, or sampling variants.
While often effective in practice, such methods operate at the symbolic level
and are inherently local: they do not directly characterize the evolving
\emph{internal state} that gives rise to long-range behaviors. As a result, they
can mitigate symptoms but offer limited insight into why collapse emerges
systematically in certain regimes, nor do they provide a principled handle for
controlling long-horizon dynamics.

We propose a geometric, dynamical-systems view of this phenomenon. We model
autoregressive decoding as a stochastic trajectory in a high-dimensional state
space (e.g., the Transformer KV cache). In this picture, mode collapse is
associated with \emph{geometric collapse}: the internal trajectory loses
\emph{state-space accessibility} and becomes confined to a low-dimensional
metastable region that is difficult to escape. This reframes the central
question from ``which tokens repeat'' to ``why does the internal dynamics become
effectively low-dimensional,'' and it suggests that purely symbolic constraints
or probability-only heuristics may be insufficient in the regimes where
collapse emerges.

We empirically support this view by measuring the \emph{correlation dimension}
of LLM generation trajectories. Correlation dimension is a trajectory-wise
fractal dimension that quantifies the dynamically active degrees of freedom.
Thus, it provides a direct geometric measure of accessibility. Across real LLM
decoding, it consistently reflects mode collapse on the text surface (including
explicit looping and increasing repetition), while remaining conceptually
distinct from token-level proxies such as entropy or Distinct-$n$.

Guided by this diagnosis, we seek to prevent mode collapse by intervening in the
model's \emph{state space} rather than only reshaping the next-token
distribution. We propose \emph{Reinforced Mode Regulation} (RMR), a lightweight
inference-time intervention that selectively attenuates the dominant
self-reinforcing directions in the Transformer value cache. RMR identifies a
small low-rank subspace exhibiting unusually strong temporal persistence by
solving a generalized eigenvalue problem with bounded spectrum. We then apply
principled thresholding to target only the most persistent directions, and
implements the regulation as a low-rank update to the value cache.

Across multiple LLMs and decoding settings, RMR substantially reduces
mode-collapse incidence, including in very low-randomness regimes where standard
decoding nearly always collapses. For instance, RMR improves non-collapse rates
from 8\% to 56\% at temperature 0.7 and from 5\% to 33\% at an entropy target of
1.0. This demonstrates that geometric, state-space regulation can stabilize
long-horizon generation without degrading text quality.

\section{Related Work}
\label{sec:related-works}

\subsection{Token-level Decoding Controls and Degeneration}
The notion of \emph{degeneration} in neural text generation was popularized by
\citet{holtzman2020curious}, which highlighted that the most probable token is
not always the best continuation. A wide range of decoding heuristics seek to
improve generation by reshaping the next-token distribution (e.g., truncation,
penalties, and sampling variants). Locally typical sampling
\citep{meister2023locally} selects tokens whose surprisal is close to the local
entropy, aiming to avoid both overly likely and overly unlikely choices. These
methods operate at the token/probability level; our focus is complementary: we
study the internal dynamics underlying mode collapse and propose a state-space
intervention. The gap between token-level, microscopic decoding and long-range,
macroscopic generation behavior is shared by many other fields, such as
LLM hallucination studies \citep{huang2025survey}.

\subsection{Repetition and Mode Collapse in Text Generation}
Prior work has proposed diverse explanations for repetition and looping in
autoregressive decoding. Some accounts attribute repetition to structural
properties of token-transition probabilities (high-inflow patterns)
\citep{fu2021theoretical} or to self-reinforcing feedback in next-token
prediction \citep{xu2022learning}. Others emphasize data-driven origins of
repetition in training corpora \citep{li2023repetition}, or propose
pattern-detection and distribution-adjustment schemes to avoid structural
repetitions in constrained domains such as code \citep{dong2025rethinking}. Our
work provides a different perspective: mode collapse is associated with
geometric collapse of the internal trajectory, related to the dynamical-system
aspects of the LLM.

\subsection{Geometric Diagnostics of Internal Dynamics}
Recent work has explored geometric characterizations of language and LLM
representations, including fractal-
\citep{doxas2010dimensionality,alabdulmohsin2024fractal,dt2025corrdim} and
intrinsic-dimension perspectives
\citep{campadelli2015intrinsic,aghajanyan2021intrinsic}. More broadly,
long-range statistical diagnostics such as scaling laws and repeated-subsequence
entropy have been used to evaluate whether generated text preserves
natural-language structure beyond local fluency
\citep{takahashi2019evaluating,tanakaishii2026repeated}. In parallel,
mechanistic interpretability research has developed tools for reading out
intermediate model beliefs, such as the tuned lens \citep{belrose2023tuned},
which decodes layerwise latent predictions. These approaches share the goal of
connecting observable behavior to internal structure; our contribution is to
provide a dynamical-systems perspective and use correlation dimension as a
trajectory-level diagnostic of state-space accessibility and relate it directly
to mode collapse in decoding.

\subsection{Inference-time Interventions}
RMR is most closely related to inference-time activation interventions
\cite{zou2023representation, turner2023steering, meng2022memit}, but differs in
goal and mechanism: we target persistent, self-reinforcing directions identified
by a bounded-spectrum generalized eigenvalue problem, and regulate the value
cache to reduce mode collapse. In addition, we study the long-range behavior of
LLM from a dynamical-systems perspective, rather than for a specific task.

\begin{figure*}
	\centering
	\includegraphics[width=\linewidth]{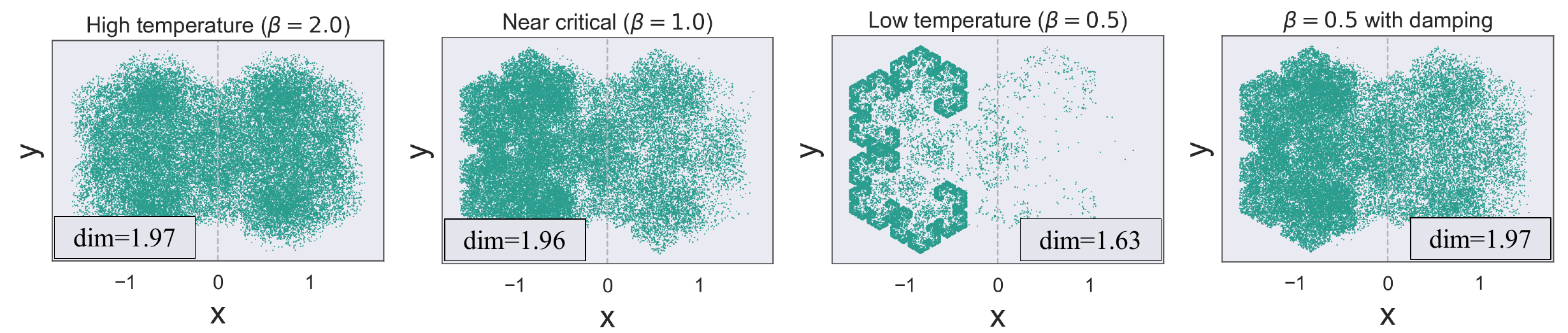}

	\caption{Trajectories of the state-dependent IFS in Eq.~\eqref{eq:ifs}. We use
    $r_i=0.6$ ($\forall i$), $\mathbf{b}_i=[2g_i,0]^\top$, and a rotation
    $\mathbf{O}_i$ with angle $\theta_i=-g_i\pi/3$.
    (a--c) Different inverse temperatures $\beta$: (a) $\beta=2.0$, (b) $\beta=1.0$,
    (c) $\beta=0.5$.
    (d) Regulation implemented as weak damping
    (Eq.~\eqref{eq:toy-damping}) applied
    to the history-dependent variable $m_t$.}
	\label{fig:ifs}
\end{figure*}

\section{The Core Idea}\label{sec:idea} 

We model autoregressive generation as an evolution of an internal state, and our
core hypothesis is that mode collapse is a symptom of an underlying {\em
geometric collapse}: during decoding, the internal trajectory becomes confined
to a low-dimensional region of representation space, i.e., state-space
accessibility collapses. A canonical and easily observable manifestation is
\emph{explicit looping}, where the model enters a periodic regime and repeatedly
produces the same token or short token sequence. See examples in
Appendix~\ref{app:examples}. This view shifts the focus from token-level
probabilities to the geometry of the underlying dynamics, and motivates
intervening directly in the state evolution.

We use a minimal dynamical model to illustrate how such accessibility collapse
can arise and how a simple regulation principle can prevent it. We then verify
in real LLM decoding that geometric collapse (measured by correlation dimension)
is closely associated with symbolic mode collapse (Section~\ref{sec:access}),
and finally introduce an LLM-specific method (Section~\ref{sec:method}).

\subsection{A Minimal Model}\label{sub:minimal-model} 

We present a minimal dynamical model to illustrate the geometric origin of LLM
looping. The model is based on an {\em iterated function system} (IFS)
\cite{hutchinson1981fractals,barnsley1985iterated,feng2009dimension}, augmented
with state-dependent map selection to capture temperature-induced phase
transitions. We refer to this construction as a \emph{state-dependent IFS}.
Figure~\ref{fig:ifs}(a–c) visualizes the system’s long-term behavior under
different temperatures.

For clarity, we consider a two-dimensional state space $X=\mathbb{R}^2$. The
system consists of $2m$ contraction maps
$\{f_i:\mathbb{R}^2\to\mathbb{R}^2\}_{i=1}^{2m}$. The first $m$ maps contract
the trajectory toward the left half-plane, while the remaining $m$ maps
contract it toward the right. Let $g_i\in\{-1, +1\}$ denote the group that map
$i$ belongs to, setting $g_1=\ldots=g_m=-1$, $g_{m+1}=\ldots=g_{2m}=1$.

Let $\mathbf{x}_t=(x_t,y_t)$ denote the system state at time $t$, and let
$\pi_t$ be a time-dependent distribution over the maps. The system evolves as
\begin{align}
\mathbf{x}_{t+1} &= f_i(\mathbf{x}_t)
= r_i \mathbf{O}_i \mathbf{x}_t + \mathbf{b}_i,
\qquad i \sim \pi_t,
\label{eq:model}
\\
\pi_t(i) &=
\exp \left(\frac{m_t g_i}{\beta} \right) \Bigm/ {\sum_{j=1}^{2m} \exp\left( \frac{m_t g_j}{\beta} \right)},
~~~~~\forall i,
\nonumber
\end{align}
where $0<r_i<1$ is the contraction factor, $\mathbf{O}_i$ is a rotation matrix,
and $\mathbf{b}_i$ is a translation vector. The quantity $m_t$ denotes a
history-dependent feedback term defined as
\begin{equation}
    m_t = \frac{1}{t}\sum_{s=1}^t x_s,
    \label{eq:order-parameter}
\end{equation}
representing the cumulative bias of the trajectory along the horizontal axis.
$g_i$ decides the direction of the bias that points to either group of maps.

When the historical average $m_t$ becomes negative, probability mass in $\pi_t$
shifts toward maps contracting into the left half-plane, which further
reinforces the trajectory bias. This positive feedback mechanism is analogous to
the mean-field Ising model for explaining the phase transition in magnetic
materials. There exists a critical temperature $\beta_0>0$ such that when
$\beta>\beta_0$, the system admits a unique ergodic invariant measure (i.e., a
single connected component), whereas for $\beta<\beta_0$, two stable long-run
regimes emerge, corresponding to trajectories confined to the left or right
half-plane.

Figures~\ref{fig:ifs}(a-c) illustrate this behavior. At high temperature (a-b),
the trajectory explores the full state space and sufficiently fills the ambient
2D space. Below the critical temperature (c), the trajectory becomes trapped in
a local attractor associated with a sub-IFS. This transition provides a minimal
geometric explanation for accessibility collapse: the accessible region shrinks
and the fractal dimension, introduced in
Section~\ref{sub:fractal-dimension}, becomes lower in (c) than in (a-b). The
resulting trajectory is more regular and predictable, analogous to mode collapse
behaviors in LLM generation, such as repetition and looping. Although LLMs do
not strictly satisfy contraction conditions, this model captures the essential
near-looping regime where effective trajectories contract toward local,
low-dimensional ``attractors.''

This model is not intended as a faithful derivation of Transformer KV-cache
dynamics. Rather, it is a deliberately restricted system that isolates one
mechanism shared with autoregressive decoding: the state produced by previous
steps influences future transition probabilities, allowing small historical
biases to become self-reinforcing over long horizons. We therefore use it only
as a qualitative model of accessibility collapse and critical transition; the
LLM-specific mechanism is studied empirically in Sections~\ref{sec:access}
and~\ref{sec:method}.

\subsection{Restoring State-Space Accessibility via Regulation}
\label{sub:damping} 

In the minimal model, accessibility collapse arises from the accumulation of
the history-dependent bias term $m_t$, which becomes dynamically dominant only
in the low-temperature regime. Suppressing this accumulation can therefore
prevent geometric collapse even when the temperature is low.

Concretely, we introduce a weak regulation factor $\eta>0$ (which can be viewed
as damping),
\begin{equation}
m_t \leftarrow (1-\eta)\; m_t,
\label{eq:toy-damping}
\end{equation}
at every timestep $t$, which limits long-term bias accumulation while preserving
short-term dynamics. Figure~\ref{fig:ifs}(d) shows that even a very small
regulation strength ($\eta=10^{-4}$) suffices to restore state-space
accessibility at low temperature.

From a dynamical perspective, variables such as $m_t$ represent directions of
persistent temporal influence. Near critical regimes, these directions evolve
significantly more slowly than the fast-mixing components of the system, and
therefore dominate long-term behavior. Such emergent persistence has been
studied across multiple theoretical frameworks, including nonequilibrium
statistical mechanics \cite{mori1965transport,zwanzig1960ensemble,te2019mori,chorin2000optimal,hohenberg1977theory},
dynamical-systems and multiscale analysis \cite{carr2012applications,fenichel1979geometric,kuehn2011mathematical},
and synergetics/self-organization \cite{haken1977synergetics}.
Importantly, slow behavior here is not assumed a priori, but emerges as a
consequence of geometric collapse.

While this form of regulation is straightforward in low-dimensional systems, identifying
persistent directions from high-dimension observations remains challenging.
In the following section, we introduce a practical method to detect and
regulate such geometric collapse during LLM decoding.

\section{Geometric Collapse in LLM Decoding}\label{sec:access} 

In this section, we move from the abstract core idea to empirical evidence in
real LLM decoding. We start from a canonical symbolic manifestation of mode
collapse---explicit looping---and then show how the correlation dimension
captures an underlying geometric collapse of the internal trajectory that
accompanies such failures.

\begin{figure*}
    \small
    \begin{center}
    \begin{minipage}{0.42\linewidth}
        \centering
        \includegraphics[width=\linewidth]{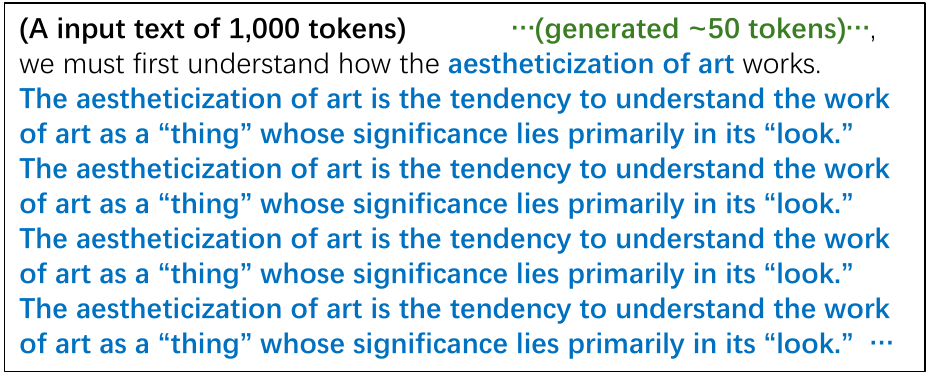} \\
        \vspace{0.8em}
        (a) Example of explicit looping
    \end{minipage}
    \hspace{0.04\linewidth}
    \begin{minipage}{0.48\linewidth}
        \centering
        \includegraphics[width=\linewidth]{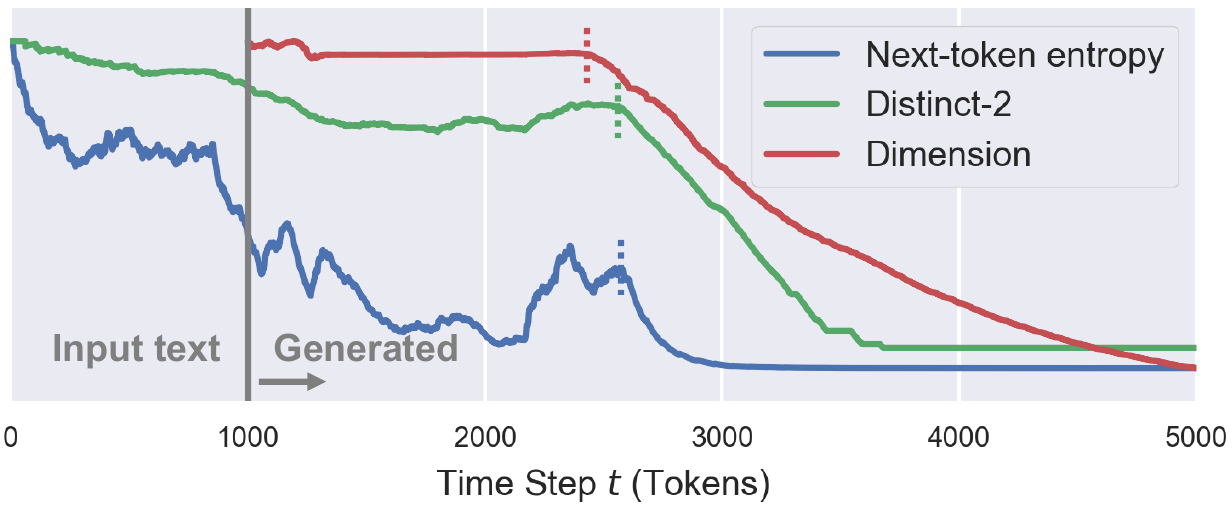} \\
        (b) Correlation dimension vs.\ token-level proxies
    \end{minipage}
    \end{center}
    \caption{
        From explicit looping to geometric collapse.
        (a) An example of explicit mode collapse (looping) with Qwen3-4B-Base at
        temperature $0.5$, chosen to make the visible token-level loop appear
        early and clearly.
        (b) A separate trajectory generated at temperature 1.0.
        Generated part starts at $t=1000$.
        Next-token entropy (blue) and Distinct-2 (green) provide token-level
        proxies of diversity, whereas correlation dimension (red) directly
        measures state-space accessibility; it drops sharply as the trajectory
        concentrates into a low-dimensional regime.
    }\label{fig:looping}
\end{figure*}

Figure~\ref{fig:looping} highlights why a geometric probe is useful: even when
token-level proxies change gradually or are sensitive to the particular
realization, the correlation dimension provides a direct and robust signal of
reduced accessibility. In intermediate regimes, generated text often already
shows degeneration---template reuse, semantic stagnation, or local variations
without new content---before entropy or Distinct-$n$ registers a sharp change.



In many nonlinear systems, long-term behavior is confined to a restricted
subset of the state space. Classical dynamical systems theory describes such
phenomena using attractors; however, for high-dimensional stochastic systems
such as large language models, attractors are generally neither well-defined
nor directly observable. Instead, what can be empirically assessed is the {\em
effective accessibility} of the state space over finite time horizons.

While the state space $X$ can be high-dimensional, a large region of the space
is often statistically inaccessible, i.e., with zero probability that a
trajectory will visit the region sufficiently often. Therefore, the notion of
accessibility can be formalized via a probability distribution (measure) $\mu$
over $X$. Moreover, $\mu$ can be {\em invariant} under $f$, and $\mu$ is called
an {\em invariant measure} of the system.

From this perspective, accessibility reflects the number of (dynamically) active
degrees of freedom that remain active during generation. Normal text generation
corresponds to trajectories that continue to explore a large, high-dimensional
region, whereas mode collapse corresponds to a progressive trapping into a much
lower-dimensional subset. A simple illustrative example is provided by the
system $(r, \theta)$ in polar coordinates: $\dot{r} = 1 - r$, $\dot{\theta} =
\text{constant} \neq 0$, whose trajectory converges to the unit circle. While
the ambient space is two-dimensional, the long-term motion is effectively
one-dimensional.

In this work, we characterize state-space accessibility using the {\em
correlation dimension}, a trajectory-wise fractal dimension originally
introduced in nonlinear time-series analysis
\cite{grassberger1983measuring,osborne1989finite} and applied to many real-world
systems \cite{small1998correlation,lacasa2013correlation,dt2024corrdim}.
Correlation dimension quantifies how the number of accessible states scales with
resolution, and thus provides a direct geometric measure of effective
dimensionality. In Section~\ref{sub:fractal-dimension}, we formally define this
quantity and show that mode collapse in large language models is closely
associated with a pronounced low-dimensional collapse of the generation
trajectory.

\subsection{Fractal Dimension}
\label{sub:fractal-dimension}

Fractal dimension generalizes the notion of dimension to non-integer values.
Non-integer dimension typically appears in fractals, which are geometric
structures that exhibit self-similarity at different scales.

The correlation dimension of a trajectory $\{x_1,x_2,\ldots\}$ is defined as the
scaling exponent $d$ in the relation
\begin{equation}
C_t(\varepsilon) \propto \varepsilon^d
\quad \text{as } \varepsilon \to 0, t\to\infty
\label{eq:corrdim}
\end{equation}
where
\begin{equation}
C_t(\varepsilon)=\frac{2}{t(t-1)} \sum_{i=1}^{t-1} \sum_{j=i+1}^t
\mathbf{1}\left(\lVert x_i-x_j\rVert < \varepsilon\right)
\label{eq:corrint}
\end{equation}
is called the correlation sum, and represents the fraction of point pairs whose
distance is smaller than $\varepsilon$. In practice, the dimension is estimated
from the slope of $\log C_t(\varepsilon)$ versus $\log \varepsilon$ over a
finite scale range; a more rigorous discussion is provided in
Appendix~\ref{app:fractal}.

To analyze the evolution of generation dynamics, we compute correlation
dimension online at each timestep. Specifically, we define the
\emph{finite-time correlation dimension} as follows.

\begin{definition}[Finite-Time Correlation Dimension]
Given a trajectory $x_1,\ldots,x_t$, the finite-time correlation dimension
$d_t^{(\varepsilon_0,\varepsilon_1)}$ is defined as the best-fit slope of
$\log C_t(\varepsilon)$ versus $\log \varepsilon$ for $\varepsilon$ sampled
log-uniformly from $(\varepsilon_0,\varepsilon_1)$.
\label{def:ftcd}
\end{definition}

Although the naive computation of $C_t$ requires $O(t^2)$ time, it admits an
efficient online update:
{
\footnotesize
\begin{equation*}
C_{t+1}(\varepsilon)
= \frac{t-1}{t+1} C_t(\varepsilon)
+ \frac{2}{t(t+1)} \sum_{i=1}^{t}
\mathbf{1}\left(\lVert x_i-x_{t+1}\rVert<\varepsilon\right),
\label{eq:ftcd}
\end{equation*}
}
which allows $d_t$ to be computed in $O(t)$ time.

For LLMs, we used the next-token log-probability vector as dynamical-system
state $x_t$, and measured the correlation dimension using the log-probability
vector sequence. Detailed settings are provided in Appendix~\ref{app:corrdim}.

\subsection{Geometric Collapse in LLM Generation}

Explicit looping (Figure~\ref{fig:looping}(a)) is a representative manifestation
of geometric collapse in LLM, corresponding to extremely limited state-space
accessibility and thus exceptionally low correlation dimension.

Previous studies often assess looping with symbolic quantities like next-token
conditional entropy and Distinct-$n$ (i.e., proportion of distinct $n$-grams in
the generation sequence), but both are sensitive to intrinsic linguistic
variation.

In Figure~\ref{fig:looping}(b), entropy fluctuates well before looping and
Distinct-2 drops only after repetition becomes explicit. In contrast,
correlation dimension remains stable during normal generation and decreases only
when the trajectory exhibits geometric collapse.

\begin{figure*}
    \small
    \begin{center}
    \begin{minipage}{0.48\linewidth}
        \centering
        \begin{minipage}{0.49\linewidth}
            \centering
            \includegraphics[width=\linewidth]{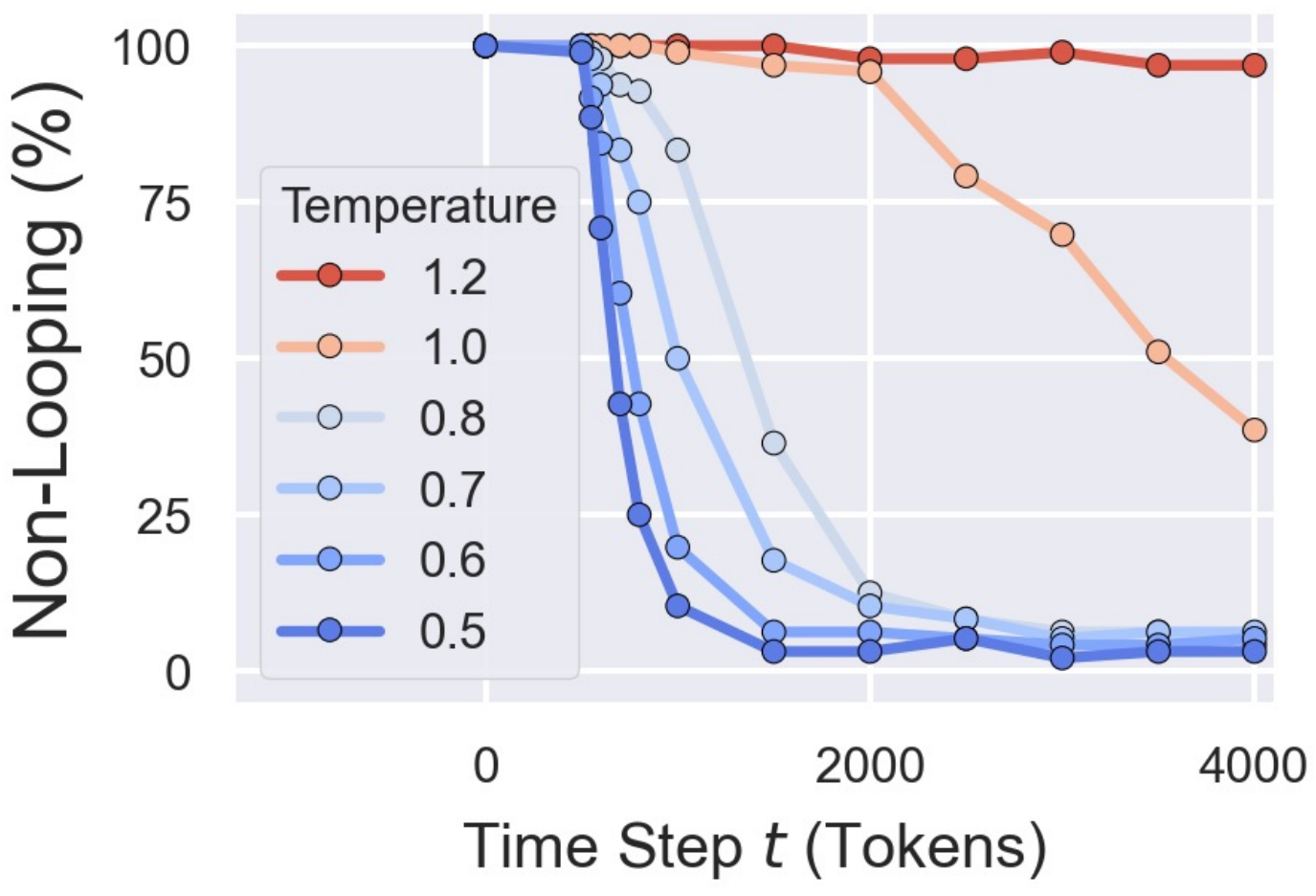} \\
            (a) Looping rate vs. temp.
        \end{minipage}
        \begin{minipage}{0.49\linewidth}
            \centering
            \includegraphics[width=\linewidth]{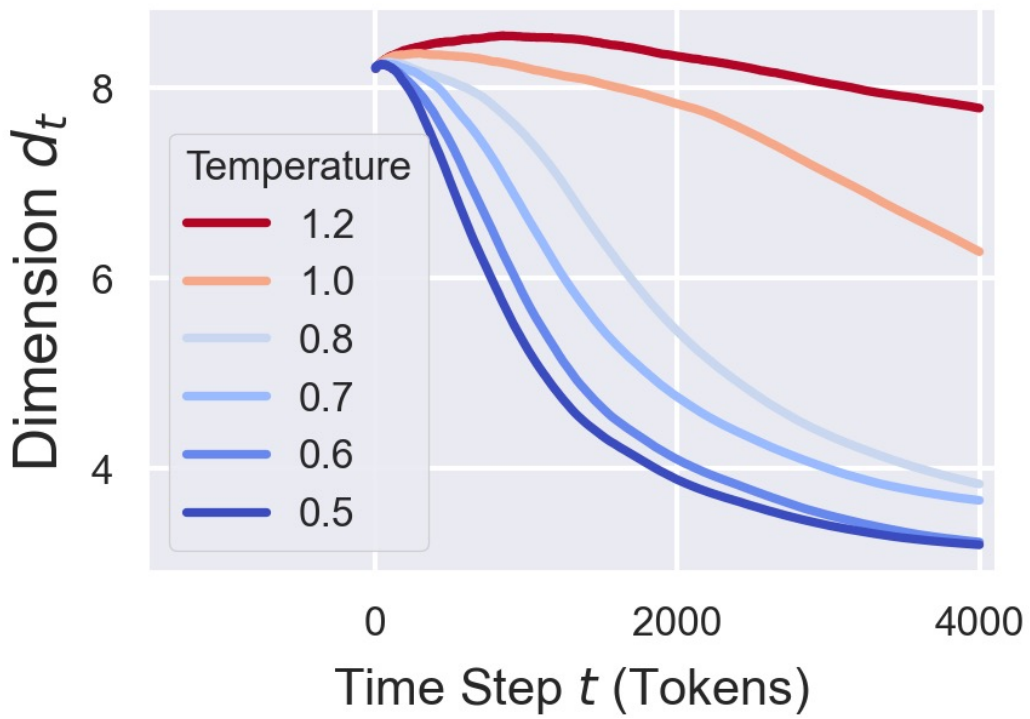} \\
            (b) Dimension vs. temp.
        \end{minipage}\\
        \vspace{0.5em}
        Temperature-locked decoding
    \end{minipage}
    \vspace{0.02\linewidth}
    \begin{minipage}{0.48\linewidth}
        \centering
        \begin{minipage}{0.49\linewidth}
            \centering
            \includegraphics[width=\linewidth]{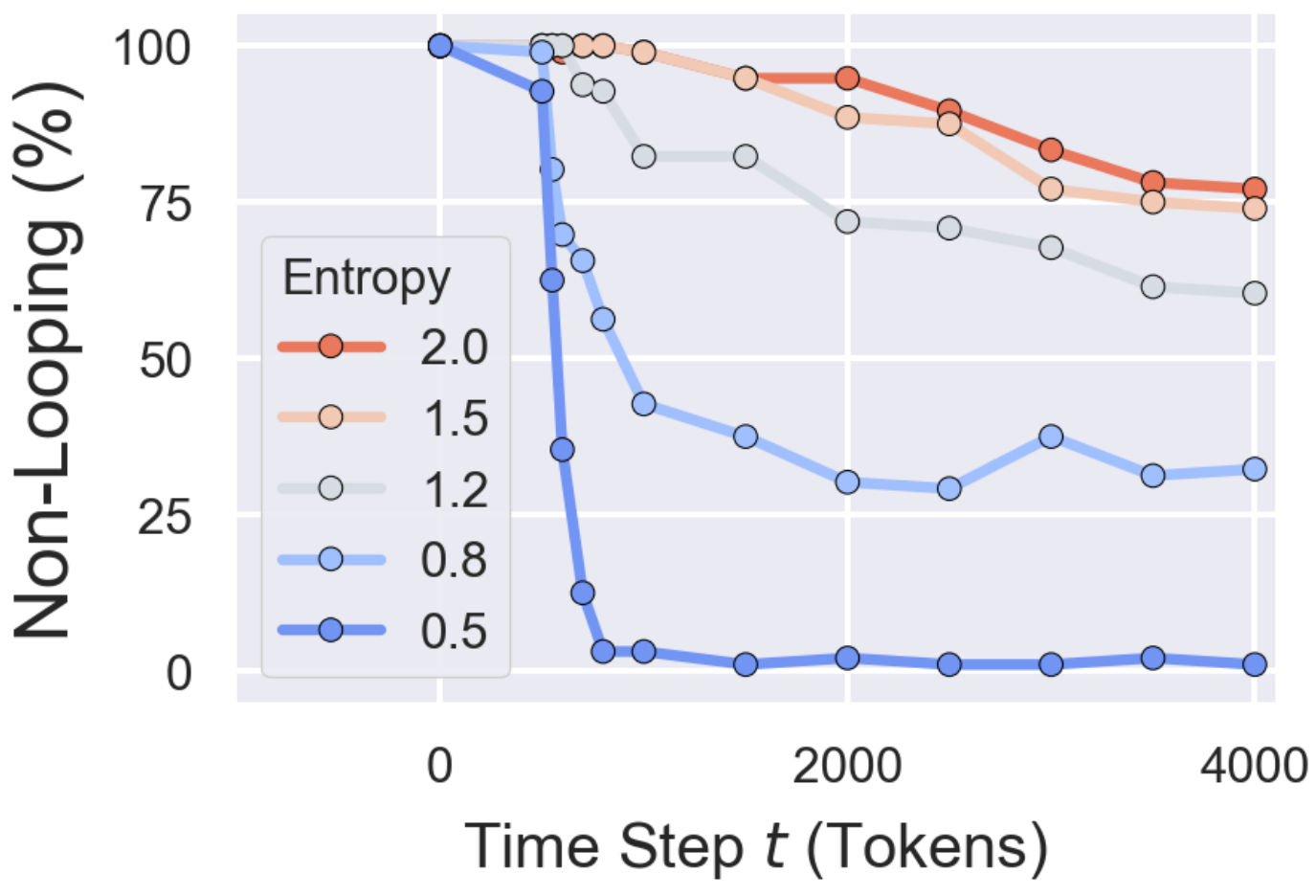} \\
            (c) Looping rate vs. entropy
        \end{minipage}
        \begin{minipage}{0.49\linewidth}
            \centering
            \includegraphics[width=\linewidth]{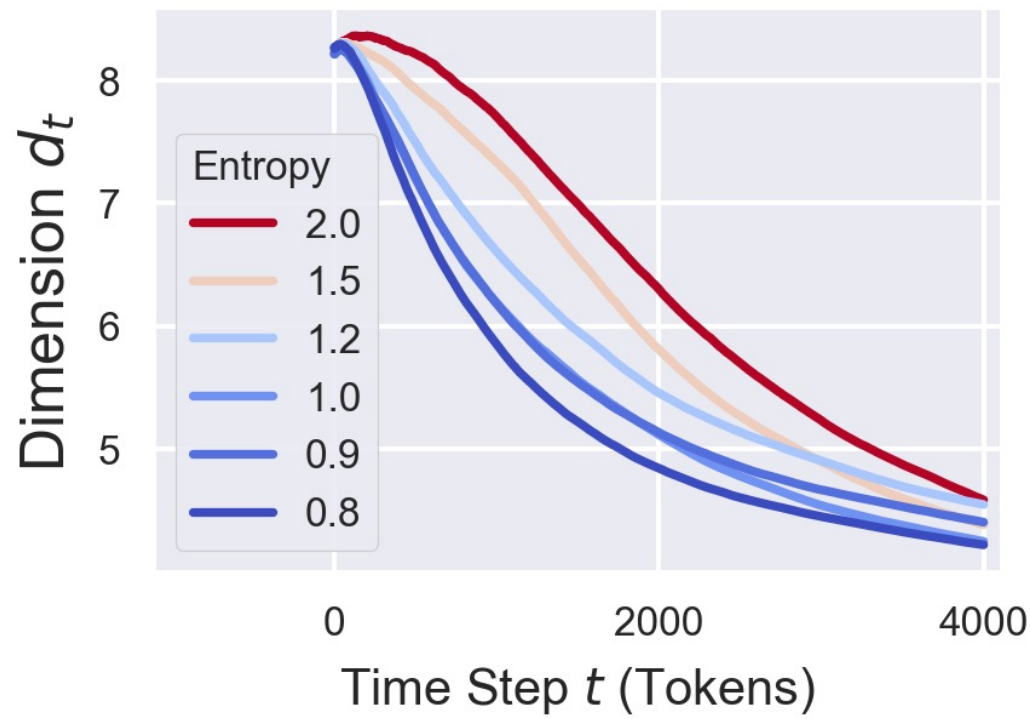} \\
            (d) Dimension vs. entropy
        \end{minipage} \\
        \vspace{0.5em}
        Entropy-locked decoding
    \end{minipage}
    \end{center}
    \caption{
        Correlation dimension tracks mode collapse across decoding conditions.
        As randomness is reduced (lower temperature or lower entropy target),
        explicit looping becomes more frequent, and correlation dimension
        decreases in tandem, indicating progressive concentration into
        low-dimensional regimes.
    }\label{fig:corrdim}
\end{figure*}

Many studies have observed that lowering generation temperature increases the
risk of looping \cite{nakaishi2024critical,pipis2025wait}, and
Figures~\ref{fig:corrdim}(a-b) verify this observation and show a relationship
between looping rate and the temperature. Similar results are observed in
Figure~\ref{fig:corrdim}(c-d) where the temperature is adaptively adjusted while
maintaining a constant next-token entropy. In (a) and (c), mode collapse is
measured by the exact looping rate, while in (b) and (d), we report the mean
correlation dimension over multiple generation runs. A monotonic decrease in
dimension is observed along generation, with significantly stronger collapse at
low temperatures or low entropy constraints. A clear correspondence between
looping rate and correlation dimension is observed in all decoding conditions.

\paragraph{Limitations of symbolic explanations.}
The collapse at low entropy constraints shows limitations of the symbolic
explanations of mode collapse (see Section~\ref{sec:related-works}), that the
generation process enters a process where entropy is gradually reduced to 0. In
the entropy-locked decoding experiment, however, the collapse still occurs if
the entropy constraint goes below a crticial value.

On the other hand, even if the generation entropy is set to a relatively high
constant value (e.g., 2.0) when exact looping is less observed
(Figure~\ref{fig:corrdim}(c)), geometric collapse still occurs (as in (d)),
manifesting in more implicit forms, such as template repetition or semantic
looping. This phenomenon corresponds to what has been termed \emph{degeneration}
in \citet{holtzman2020curious}, which includes more subtle failures such as
incoherence or blandness. Representative examples of this softer form of
collapse are provided in Appendix~\ref{app:examples}.

This indicates the unique value of correlation dimension in reflecting mode
collapse in LLM generation. Different from symbolic quantities like entropy or
Distinct-$n$, correlation dimension captures the active degrees of freedom in a
generation trajectory.

\section{Reinforced Mode Regulation for LLM}
\label{sec:method}

As demonstrated in Sections~\ref{sec:idea}, geometric collapse is a key
mechanism underlying mode collapse, and regulating a small number of persistent
components can effectively prevent such collapse.

In this section, we propose \emph{Reinforced Mode Regulation} (RMR), a
lightweight state-space intervention for Transformer-based LLMs. RMR regulates a
small set of directions in the value-cache space that exhibit unusually strong
temporal persistence, thereby preventing the cache trajectory from concentrating
along a single persistent direction. Concretely, RMR can be implemented as a
low-rank damping operator applied to the value cache along these directions.
These directions are identified by a generalized eigenvalue problem whose
eigenvalues admit a natural, bounded scale; the eigenvalue directly quantifies
persistence and enables principled thresholding (damping).

\begin{figure*}
    \small
    \begin{center}
        \begin{minipage}{0.48\linewidth}
        \centering
        \includegraphics[width=\linewidth]{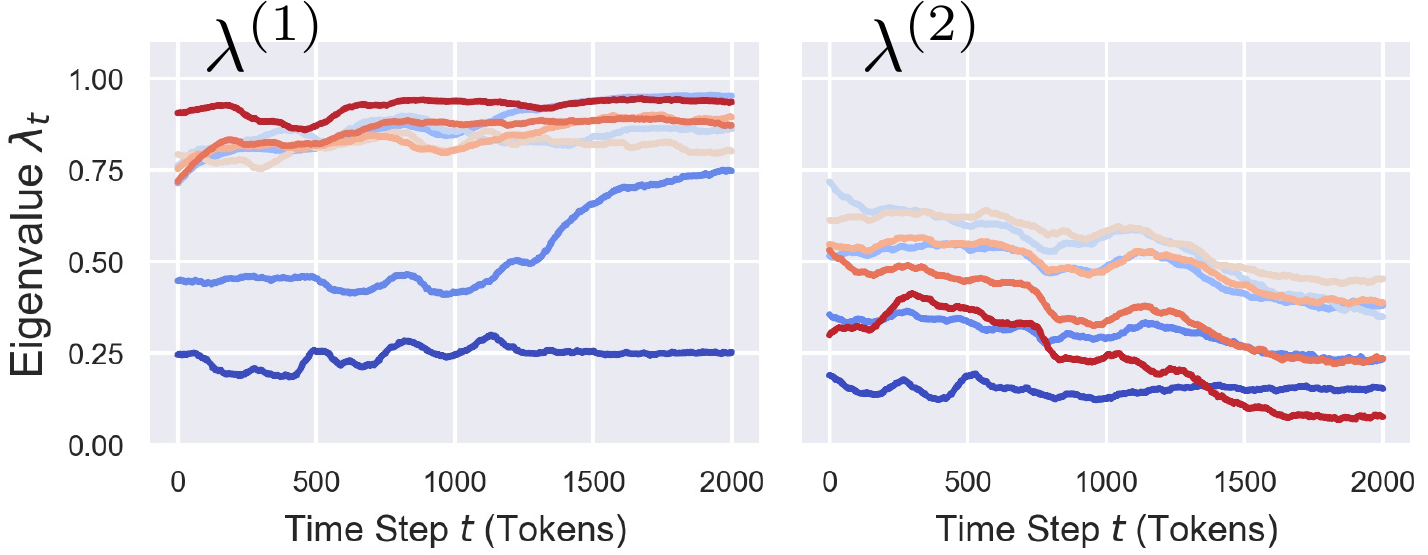}\\
        (a) RMR OFF: geometric collapse begins around $t\approx 1000$
        \end{minipage}
        \vspace{0.02\linewidth}
        \begin{minipage}{0.48\linewidth}
        \centering
        \includegraphics[width=\linewidth]{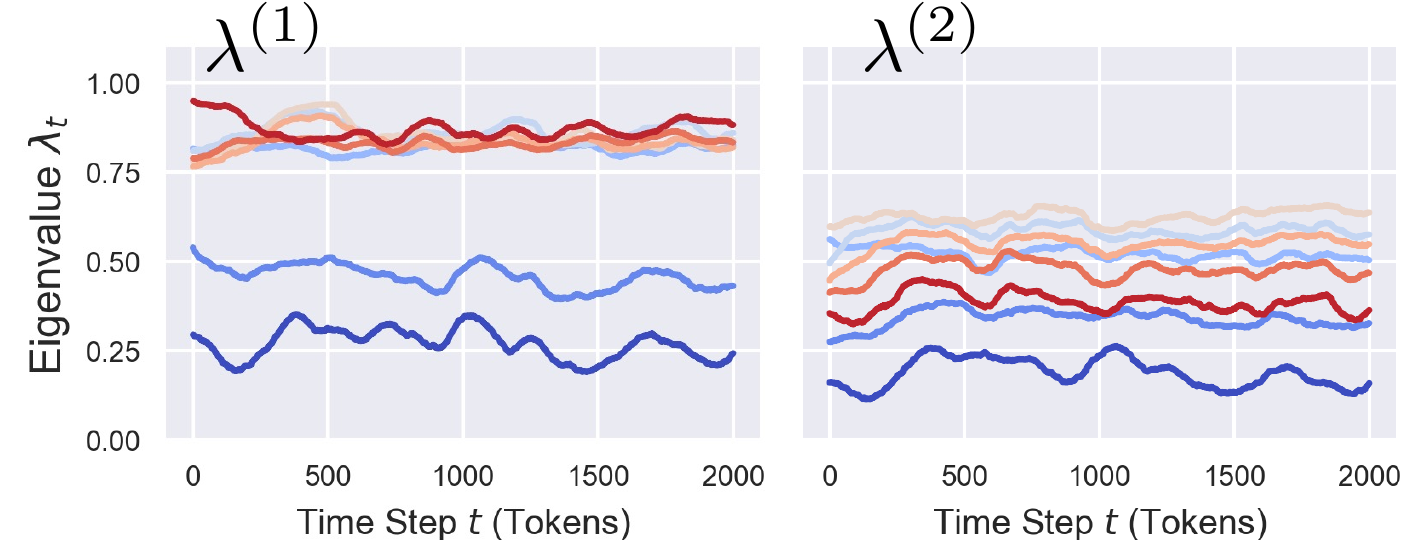}\\
        (b) RMR ON: geometric collapse is substantially mitigated
        \end{minipage}
    \end{center}
    \caption{
        Evolution of the top-2 generalized eigenvalues in the value cache across
        Transformer layers (color-coded from blue to red), reporting the mean
        over attention heads.
        (a) Without RMR, geometric collapse begins around $t\approx1000$: the
        leading eigenvalue rises toward 1 while the second eigenvalue quickly
        drops toward 0, indicating a widening spectral gap and dominance by a
        single persistent mode.
        (b) With RMR, the leading eigenvalue in each layer is regulated below
        the threshold ($\lambda_\text{min}=0.8$), maintaining a stable spectrum
        and preventing collapse.
        }\label{fig:spectral}
\end{figure*}

\begin{figure*}[t]
\begin{center}
    \small
    \begin{minipage}{0.4\linewidth}
        \centering \includegraphics[width=\linewidth]{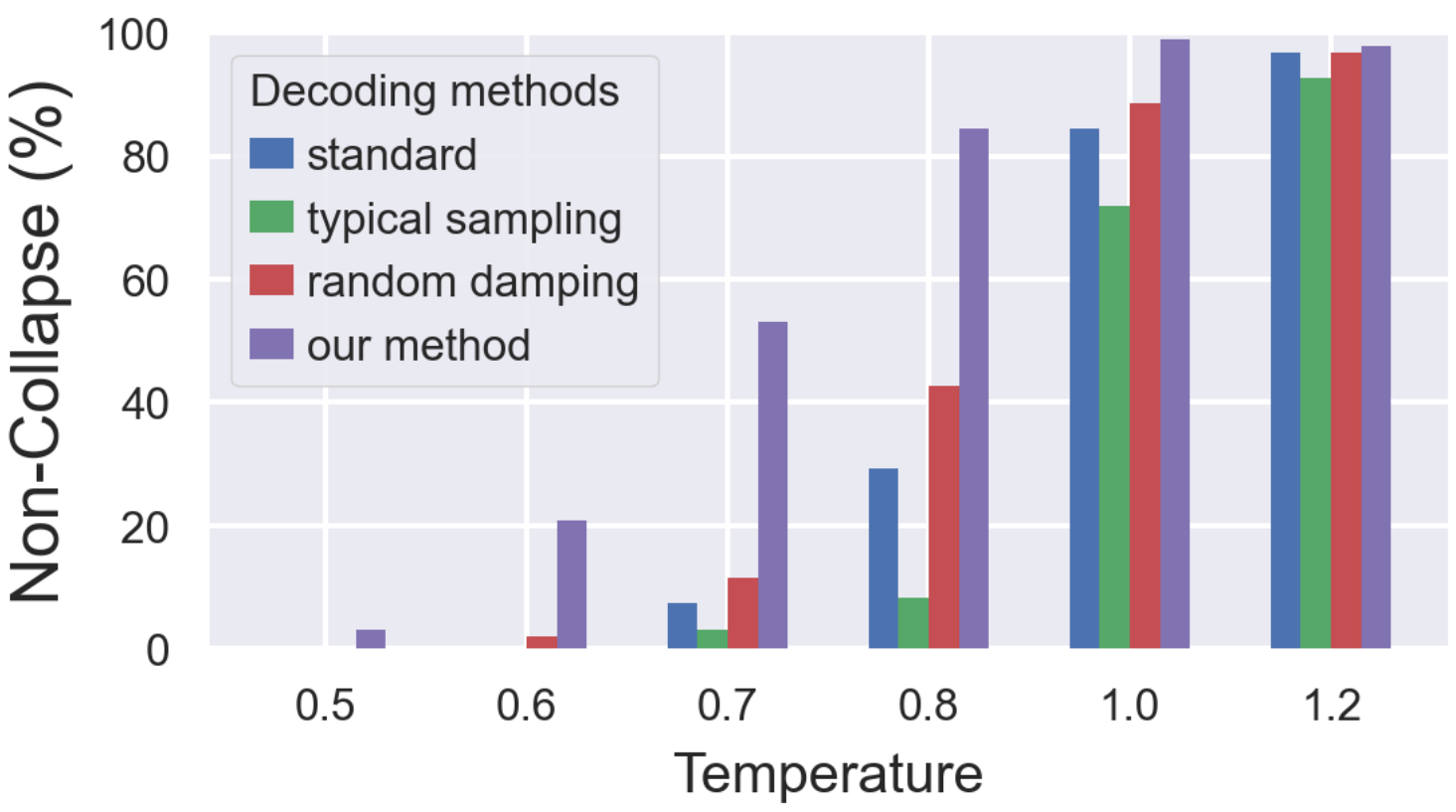}
        \\
        (a) Temperature-locked
    \end{minipage}
    \hspace{0.05\linewidth}
    \begin{minipage}{0.4\linewidth}
        \centering
        \includegraphics[width=\linewidth]{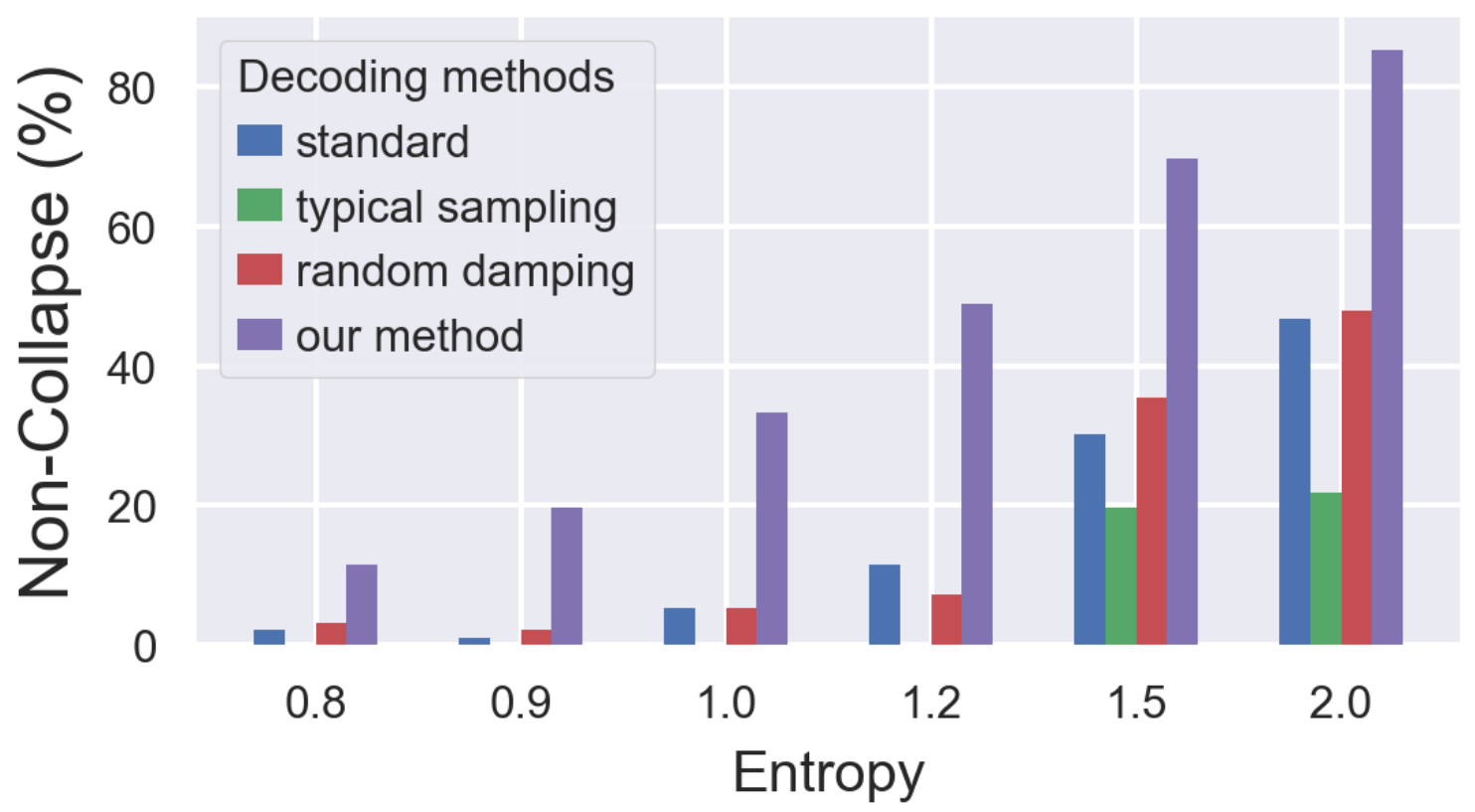} \\
        (b) Entropy-locked
    \end{minipage}
\end{center}
\vskip-0.5em
\caption{
Non-collapse rates under controlled randomness.
(a) Temperature-locked decoding and (b) entropy-locked decoding, comparing
standard decoding, typical sampling, random regulation, and RMR. A completion
is counted as non-collapse if its correlation dimension remains above 8 over
1{,}000 generated tokens of \textsc{Heidegger}.}\label{fig:noncollapse}
\end{figure*}

\paragraph{Key-Value Cache as State of Transformer}
In a Transformer, the generation state at timestep $t$ is fully represented by
the key-value (KV) cache across layers,
\begin{equation}
\{(\mathbf{K}_t^{(l)}, \mathbf{V}_t^{(l)})\}_{l=1}^L,
\qquad
\mathbf{K}_t^{(l)}, \mathbf{V}_t^{(l)} \in \mathbb{R}^{t \times D},
\end{equation}
where batch and head dimensions are omitted for clarity.

At each decoding step, the cache is extended as
\begin{equation}
\mathbf{K}_t = [\mathbf{K}_{t-1}; \mathbf{k}_t],
\qquad
\mathbf{V}_t = [\mathbf{V}_{t-1}; \mathbf{v}_t],
\end{equation}
and the value cache is aggregated through attention:
\begin{equation}
\text{Attention}(\mathbf{q}_t;\mathbf{K}_t,\mathbf{V}_t)
=
\text{softmax}\!\left(
\frac{\mathbf{q}_t \mathbf{K}_t^\top}{\sqrt{D}}
\right)\mathbf{V}_t.
\end{equation}

We focus on the value-cache matrices $\mathbf{V}_t$. The discussion below
considers an arbitrary layer and applies analogously to other layers. We do
{\em not} regulate the key-cache matrices, since modifying keys changes
attention weights and can disrupt the global temporal dependency structure.

We view $\mathbf{V}_t \in \mathbb{R}^{t \times D}$ as a sequence of vectors
collected along the temporal dimension. When generation approaches a
degenerative regime, we assume that geometric collapse concentrates along a
low-dimensional subspace, represented by an orthonormal basis
$\mathbf{U}\in\mathbb{R}^{D\times c}$. Damping is then applied selectively to
the projection onto this subspace:
\begin{equation}
\mathbf{V}_t
\leftarrow
\mathbf{V}_t-\eta\mathbf{\Gamma}\mathbf{V}_t
\mathbf{U}\mathbf{U}^\top,
\label{eq:damping}
\end{equation}
where $\eta>0$ controls the damping strength and
$\mathbf{\Gamma}=\mathrm{diag}(\gamma^t,\gamma^{t-1},\ldots,\gamma^1)$ applies
exponentially weaker damping to older timesteps, with $\gamma$ set to 0.995
in our experiments.

To identify the dominant persistent directions, we estimate $\mathbf{U}$ by
solving the generalized eigenvalue problem
\begin{equation}
\mathbf{\Sigma}_\Delta \mathbf{u}
=
\lambda\,\mathbf{\Sigma}\mathbf{u},
\label{eq:slow}
\end{equation}
where
\begin{align}
\mathbf{\Sigma} &=
\mathbb{E}_t\!\left[
(\mathbf{v}_t-\bar{\mathbf{v}})
(\mathbf{v}_t-\bar{\mathbf{v}})^\top
\right], \\
\mathbf{\Sigma}_\Delta &=
\mathbb{E}_t\!\Bigl[
\operatorname{sym}\bigl(
(\mathbf{v}_{t+1}-\bar{\mathbf{v}})(\mathbf{v}_t-\bar{\mathbf{v}})^\top
\bigr)
\Bigr],
\end{align}
where $\operatorname{sym}(\mathbf{A}) := \tfrac{1}{2}(\mathbf{A}+\mathbf{A}^\top)$,
with $\mathbf{v}_t\in\mathbb{R}^D$ denoting the $t$-th row of $\mathbf{V}_t$ and
$\bar{\mathbf{v}}$ its temporal mean.

Under stationary conditions, the spectrum is bounded with $|\lambda|\le 1$.
Large eigenvalues ($\lambda\approx1$) correspond to directions with slow
temporal decay, indicating strong persistence. We select the top-$c$
eigenvectors to form $\mathbf{U}$. The required statistics are estimated online
using exponentially weighted moving averages (decay rate $\gamma=0.99$), and the
top eigenvectors are efficiently computed via power iteration
(Appendix~\ref{app:power-iteration}). The eigenvalues can be understood as the
{\em auto-correlation} coefficient of $\mathbf{v}_t$ along different directions.
Therefore, our method can be naturally generalized to other LLM architectures,
such as state-space models \citep{gu2024mamba}.

\paragraph{Eigenvalue Threshold and Regulation}
Thresholding is crucial for stable intervention: because Eq.~\eqref{eq:slow}
induces a bounded spectrum ($|\lambda| \le 1$ under stationarity), any threshold
has a consistent meaning across models, layers, and timesteps. We set an
eigenvalue threshold $\lambda_\text{min}=0.8$ and define $c$ as the number of
eigenvalues with $\lambda>\lambda_\text{min}$. In practice, most eigenvalues are
small, so we estimate only the top-8 eigenvalues/eigenvectors and regulate those
exceeding $\lambda_\text{min}$. For the value-cache component along the selected
subspace, we use a fixed regulation strength $\eta=0.7$ (implemented as damping
in Eq.~\eqref{eq:damping}). Compared to fixing $c$, thresholding is more stable:
even if regulation is applied at consecutive steps, ``fast'' directions with
small $\lambda$ remain unaffected, allowing mid-level persistent variables
(e.g., topic and style) to be preserved.

In practice, we attenuate only directions whose eigenvalues exceed the
threshold and reduce them to slightly below $\lambda_\text{min}$. This
eigenvalue threshold has a concrete meaning (due to the bounded spectrum) and
is therefore more reliable than regulating a fixed number $c$ of directions:
it avoids unnecessarily suppressing moderately persistent components and
reduces unintended side effects. Empirically, performance is relatively
insensitive to the regulation interval; applying damping once every 10 decoding
steps already works well in our experiments.

\paragraph{Spectral Structure and Effects of Regulation}
Figure~\ref{fig:spectral} shows the evolution of the top-2 generalized
eigenvalues (Eq.~\eqref{eq:slow}) across Transformer layers, averaged over all
attention heads (see Appendix~\ref{app:spectral} for additional details,
including both mean and maximum over heads). In
Figure~\ref{fig:spectral}(a) (RMR off), the largest eigenvalue stays close to
$1$ in most layers (except near the input embedding). Around the onset of mode
collapse ($t\approx1000$), the top-1 eigenvalue in some layers increases while
the top-2 eigenvalue quickly decreases toward $0$, indicating that the
dynamics become increasingly dominated by a single persistent direction (a
growing spectral gap).

Figure~\ref{fig:spectral}(b) shows a run with RMR where collapse is prevented.
Regulation keeps the top-1 eigenvalue in each layer below the threshold
$\lambda_\text{min}$, and both top-1 and top-2 eigenvalues fluctuate around
stable values throughout generation.



\section{Experimental Setup}
\label{sec:experiment}

Throughout the paper we use standard top-$k$ and top-$p$ filtering to improve
generation quality, with $k=50$ and $p=0.9$ (i.e., restricting next-token
candidates to at most 50 tokens with cumulative probability at least $0.9$). In
entropy-locked experiments, the target entropy is computed over the same
top-$k$ candidate set, so the reported entropy reflects uncertainty after the
top-$k$ restriction.

Unless otherwise noted, results in the main text are obtained from a fixed
prompt: the first 1{,}000-token segment of \textsc{Heidegger} from the SEP
dataset, using Qwen3-4B-Base \cite{yang2025qwen3}. Comprehensive results across
additional prompts and larger, instruction-tuned models are provided in
Appendix~\ref{app:results-sep} and Appendix~\ref{app:results-llm}.

In addition to RMR, we also consider {\em typical sampling}
\cite{meister2023locally} as a baseline. Typical sampling favors tokens whose
surprisal is close to the distribution's ``typical'' value (near the local
entropy), effectively filtering out both overly likely and overly unlikely
tokens before sampling. Following \citet{meister2023locally}, we set the typical
cumulative probability to 0.2.

\section{Results}
\label{sec:results}

\paragraph{Mitigating Mode Collapse via Geometric Regulation}

Figure~\ref{fig:noncollapse} compares RMR with standard decoding across
different levels of randomness. Panel (a) fixes the generation temperature and
panel (b) fixes the next-token entropy to the constants shown on the horizontal
axis.
Non-collapse rate is computed as the fraction of generated 1{,}000-token
completions of \textsc{Heidegger} for which the correlation dimension is above
8. We use this threshold as a geometric diagnostic rather than as a complete
definition of text quality; the same experimental section also reports
token-level and judge-based evaluations that do not use correlation dimension.

Across all temperature-locked and entropy-locked settings, RMR
consistently achieves substantially higher non-collapse rates than standard
decoding. The gain is most pronounced near the critical regime where baseline
decoding starts to fail (temperature $\approx0.7$--$0.8$). For example, at
temperature $0.7$, the non-collapse rate increases from $8\%$ (standard) to
$56\%$ with RMR. In the entropy-locked setting, RMR is the only
approach that maintains a sizable non-collapse rate under extremely low entropy
rates (e.g., entropy $=1.0$), where standard decoding nearly always collapses
(about $5\%$ non-collapse versus $33\%$ for RMR at entropy $=1.0$).

\paragraph{Baselines: Typical Sampling and Random Regulation}
Figure~\ref{fig:noncollapse} also includes two baselines. Typical sampling
\cite{meister2023locally} does not prevent collapse in the low-randomness regime
and can underperform standard decoding, consistent with its tendency to reduce
effective sampling uncertainty and push generation toward low-entropy collapse.
Random regulation---regulating a randomly chosen value-cache subspace with the
same rank and strength---also falls short. These results support that RMR’s
benefit comes from targeting the specific persistent subspace rather than
generic interventions.

\paragraph{External Collapse Indicators}
Although our primary non-collapse rate is based on correlation dimension, we
also report collapse indicators that do not use the proposed geometric
diagnostic. In Figure~\ref{fig:corrdim}(a,c), exact looping is measured directly
from the generated token sequence and follows the same low-randomness transition
as the dimension drop. Figure~\ref{fig:looping} contrasts correlation dimension
with next-token entropy and Distinct-2: these token-level quantities are less
temporally precise, but they provide independent evidence of the surface
degeneration associated with collapse. The examples in
Appendix~\ref{app:examples} further show that collapse can appear as template
repetition or semantic stagnation even before it becomes an exact token loop.

\paragraph{Text Quality Evaluation}
An important concern is whether intervening in the KV cache degrades the
surface quality of text when generation is {\em not} collapsing. To isolate
potential side effects, we evaluate only non-collapse completions (correlation
dimension $>8$) generated at relatively high temperatures (1.0 or 1.2), where
standard decoding is typically stable. Table~\ref{tab:llm-judge} reports
LLM-as-a-Judge scores comparing standard decoding and RMR. Across models and
criteria, the score distributions are statistically indistinguishable
(Kolmogorov-Smirnov (K-S) $p$-values well above
0.05), suggesting that regulating the eigenvalue-thresholded subspace
does not introduce noticeable changes to surface properties such as coherence,
syntax, and information progression. The full rating prompt is provided in
Appendix~\ref{app:prompt}.

\begin{table}[t]
    \small
    \caption{LLM-as-a-Judge scores for non-collapse texts. Completions are
        generated in a temperature-locked setting with (a) 1{,}000 tokens and
        (b) 4{,}000 tokens; only non-collapse samples are included. Large
        $p$-values indicate statistically indistinguishable score
        distributions (larger is better). }\label{tab:llm-judge}

    \vskip 1em
    \textbf{(a) Generations of 1{,}000 tokens} \\
    \begin{tabular*}{\linewidth}[c]{l @{\extracolsep{\fill}} cccc}
        \toprule \textbf{Method} & \textbf{standard} & \textbf{RMR} & \textbf{p-value} \\
        (Non-collapse rate) & (84\%) & (99\%) & \\
        \midrule \multicolumn{3}{l}{\textbf{Deepseek-R1-14B}} \\
        ~~Coherence & 8.26 (1.2) & 8.27 (1.0) & 0.33 \\
        ~~Syntax & 6.98 (1.1) & 7.07 (0.9) & 0.97 \\
        ~~Progress & 7.21 (1.2) & 7.14 (1.1) & 0.22 \\
        \midrule
        \multicolumn{3}{l}{\textbf{GPT-5-mini}} \\
        ~~Coherence & 6.32 (1.9) & 6.25 (2.1) & 0.55 \\
        ~~Syntax & 6.10 (1.6) & 6.07 (1.9) & 0.98 \\
        ~~Progress & 2.51 (1.8) & 2.22 (1.8) & 0.72 \\
        \bottomrule
    \end{tabular*} \\
    \vskip 2em
    \textbf{(b) Generations of 4{,}000 tokens.}
    $p$-values are computed between
    standard decoding (temp. $=1.2$) and RMR (temp. $=1.0$). \\
    \begin{tabular*}{\linewidth}[c]{l @{\extracolsep{\fill}} cccc}
        \toprule
        \textbf{Method} & \multicolumn{2}{c}{\textbf{standard}} & \multicolumn{1}{c}{\textbf{RMR}} & \textbf{p-value} \\
        \cmidrule(rl){2-3} \cmidrule(rl){4-4}
        \textbf{Temperature} & \textbf{1.0} & \textbf{1.2} & \textbf{1.0}\\
        (Non-collapse rate) & (6\%) & (55\%) & (99\%) & \\
        \midrule \multicolumn{3}{l}{\textbf{Deepseek-R1-14B}} \\
        ~~Coherence & 5.75 & 7.66 & 7.34 & 0.85 \\
        ~~Syntax    & 5.50 & 6.67 & 6.70 & 0.99 \\
        ~~Progress  & 5.03 & 6.69 & 6.36 & 0.28 \\
        \midrule
        \multicolumn{3}{l}{\textbf{GPT-5-mini}} \\
        ~~Coherence & 5.08 & 6.40 & 6.23 & 0.14 \\
        ~~Syntax    & 4.33 & 5.68 & 5.95 & 0.58 \\
        ~~Progress  & 2.00 & 2.73 & 2.78 & 0.99 \\
        \bottomrule
    \end{tabular*}
\end{table}



\section{Limitations}

Non-autoregressive LLMs (e.g., diffusion-based models) are not considered in
this work. RMR also assumes access to internal Transformer states (the KV
cache) during decoding, which may be infeasible in black-box settings and
requires integration into inference stacks. While RMR is lightweight, it adds
overhead for maintaining running statistics and estimating dominant
eigenvectors, which can matter under strict latency constraints or for very
large models.

We use a fixed eigenvalue threshold and regulation strength throughout our
experiments and find the method to be stable under this simple configuration; in
particular, the threshold $\tau=0.8$ is effective across all tested settings. The
main-text experiments use a fixed long-form continuation setting for clarity;
the appendix broadens this to additional SEP prompts and instruction-tuned
models, while broader coverage of multi-turn dialogue, RAG, code generation, and
reasoning tasks remains future work. Finally, our evaluation focuses on mode
collapse as characterized by correlation dimension and related repetition
indicators; interactions with other failure modes, safety behaviors, and
downstream task performance remain for future work.

\section{Conclusion}

We revisited mode collapse in autoregressive text generation from a
dynamical-systems perspective and linked symbolic failures to \emph{geometric
collapse} of the internal trajectory (reduced state-space accessibility). Guided
by this view, we proposed \emph{Reinforced Mode Regulation} (RMR), which
identifies and attenuates the dominant persistent subspace in the Transformer
value cache via a bounded-spectrum generalized eigenvalue formulation and
thresholding. Across multiple LLMs, RMR substantially improves non-collapse
rates across temperatures and entropy targets, and preserves surface text
quality in non-collapse regimes.

RMR is designed and evaluated as a decoding-time intervention. More broadly,
trajectory-level diagnostics such as correlation dimension may also be useful
for analyzing training outcomes, which remains a future work.

\section*{Acknowledgements}
This work is supported by JST CREST, Japan, Grant Number JPMJCR2114.

\section*{Impact Statement}

This paper presents work whose goal is to advance the field of Machine Learning.
There are many potential societal consequences of our work, none which we feel
must be specifically highlighted here.


\bibliography{main}
\bibliographystyle{icml2026}

\newpage
\appendix
\onecolumn


\section{Iterated Function Systems}
\label{app:ifs}
An iterated function system (IFS) consists of a collection of contraction maps
$\{f_i:X\to X\}_{i=1}^m$ on a metric space $X$ and a probability distribution
$\pi$ over the maps (with $\pi_i>0$). At each discrete time step, we sample an
index $i\sim\pi$ and update $\mathbf{x}_{t+1}=f_i(\mathbf{x}_t)$. The simplest
example is the \emph{self-similar} (affine) IFS:
\begin{equation}
\mathbf{x}_{t+1} = f_i(\mathbf{x}_t) = r \mathbf{x}_t + \mathbf{b}_i,
\qquad i \sim \pi,
\label{eq:ifs}
\end{equation}
where $r$ ($0<r<1$) is the contraction factor and $\mathbf{b}_i\in\mathbb{R}^n$
is a translation. Under standard assumptions, such affine IFS admit a unique
attractor and an ergodic invariant measure $\mu$ supported on it. Under mild
conditions, the fractal dimension of $\mu$ satisfies
\begin{equation}
	d(\mu) = \min\left\{\frac{h_\pi}{- \log r}, \; \dim(X)\right\},
\end{equation}
where $h_\pi$ is the Shannon entropy \cite{feng2009dimension} of $\pi$.


\section{Fractal Dimension and Correlation Dimension}
\label{app:fractal}

Fractal dimension can be defined with respect to a set (e.g., Hausdorff
dimension, box-counting dimension) or a measure (e.g., information dimension,
correlation dimension). The distinction between the two is whether the
distribution of the points in the set is considered. Measure-based fractal
dimension is typically used in the context of dynamical systems, where how
\emph{frequently} a trajectory visits a region is investigated. In many cases, the two
classes of fractal dimensions are close in value.

Below is a definition of measure-based fractal dimension.
\begin{definition}[Fractal Dimension of Measure]
	For a probability measure $\mu$ on a metric space $X$, the local fractal
	dimension of $\mu$ is defined as
	\begin{equation}
		d(\mu, \mathbf{x}) = \lim_{\varepsilon\to 0}
		\frac{\log \mu\bigl(B(\mathbf{x}, \varepsilon)\bigr)}{\log \varepsilon},
        \label{eq:fractal-dimension}
	\end{equation}
	where $B(\mathbf{x}, \varepsilon)$ is the closed ball centered at
	$\mathbf{x}$ with radius $\varepsilon$. If $d(\mu, \mathbf{x})$ is constant
	for $\mu$-almost all $\mathbf{x}\in X$, we denote the value by $d(\mu)$ and
	call it the \emph{fractal dimension} of $\mu$.
\end{definition}

\paragraph{The Grassberger--Procaccia estimator and Correlation Dimension}
In practice, $\mu$ is usually intractable and manifest only as the limit of the
empirical measure. Correlation dimension is essentially an estimator of the
measure of the ball $B(\mathbf{x}, \varepsilon)$ in
Eq.~\eqref{eq:fractal-dimension}.

Let $t$ denote the trajectory length. The Grassberger--Procaccia (GP) method
\cite{grassberger1983measuring} estimates the correlation sum
\begin{equation}
	C_t(\varepsilon)=\frac{2}{t(t-1)} \sum_{1\le i<j\le t}
	\mathbb{I}\!\left(\|\mathbf{x}_i-\mathbf{x}_j\|_2 \le \varepsilon\right),
\end{equation}
which approximates $\mu(B(\mathbf{x},\varepsilon))$ up to constants for small
$\varepsilon$ and large $t$. The (correlation) dimension is then estimated from
the slope of $\log C_t(\varepsilon)$ versus $\log \varepsilon$ over an
appropriate scaling range. This estimator is proposed by Grassberger and
Procaccia \cite{grassberger1983measuring}, and called the GP estimator by
convention. See \cite{pesin1993rigorous} for a rigorous mathematical definition
of correlation dimension.

\subsection{Estimation Details of Correlation Dimension}
\label{app:corrdim}

We estimate the correlation dimension of a generation trajectory
$\mathbf{x}_1,\ldots,\mathbf{x}_t$ by using the next-token log-probability
vector sequence as the state, following \cite{dt2025corrdim}. Specifically,
at each time step $t$ we define
\begin{equation}
    \mathbf{x}_t = \log \mathbb{P}(w_{t} | w_{t-1},\ldots,w_1), \quad w_{t} \in \mathcal{V},
\end{equation}
where $w_t$ is the $t$-th token in the generation sequence, $\mathcal{V}$ is the
vocabulary.

To reduce the computational cost due to the large vocabulary size (typically
above 100{,}000), we use a projection-based approximation inspired by
Marstrand's projection theorem \cite{marstrand1954some,falconer2004fractal}: we
first group the dimensions in $\mathcal{V}$ into $K=10{,}000$ bins using
$\Phi(i) = i~\text{mod}~K$, and add up the log-probabilities in $\mathbf{x}_t$
within each bin to get a 10{,}000-dimensional vector, and use it in replacement
of $\mathbf{x}_t$.

The finite-time correlation dimension defined in Definition \ref{def:ftcd}
specifies a scaling range $(\varepsilon_0, \varepsilon_1)$, in which the
correlation sum $C_t(\varepsilon)$ is expected to scale quasi-linearly with
$\varepsilon$. For Qwen3-4B-Base, we find the distances between timesteps to
range from about 100 to 10000. Therefore, we set $\varepsilon_0=500$ and
$\varepsilon_1=1200$, which is in the middle of the range in the log scale. This
setting differs from \cite{dt2025corrdim}, in which a smaller range at around
$(100, 500)$ is used. We found the setting in \cite{dt2025corrdim} to be very
sensitive to incremental changes in the generation trajectory, unsuitable for
temporal monitoring of correlation dimension.

\section{Additional Examples}
\label{app:examples}

\begin{figure}[h]
    \begin{minipage}{\linewidth}
        \centering
    \includegraphics[width=0.95\textwidth]{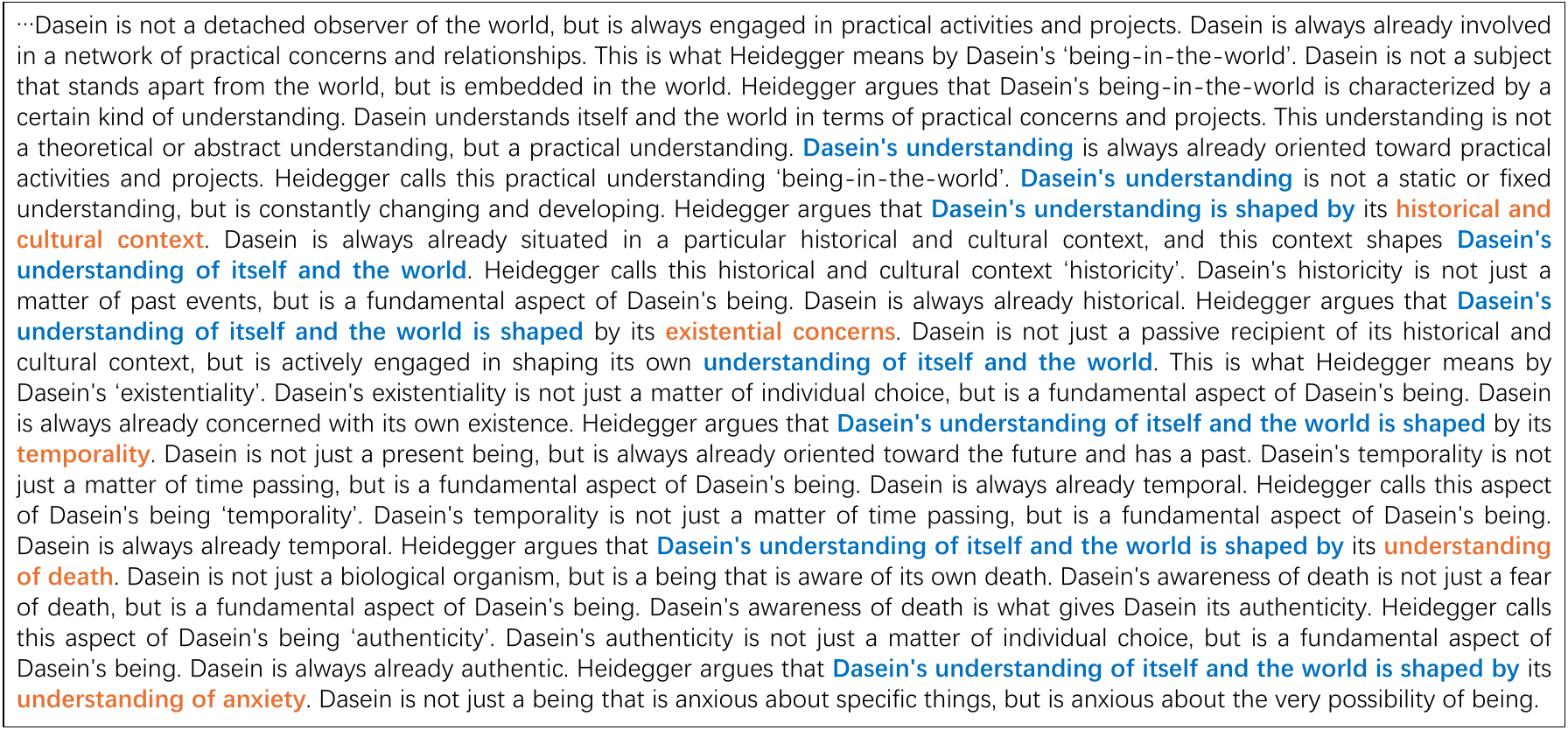}\\
    (a)
    \end{minipage}
    \begin{minipage}{\linewidth}
        \centering
        \includegraphics[width=0.95\textwidth]{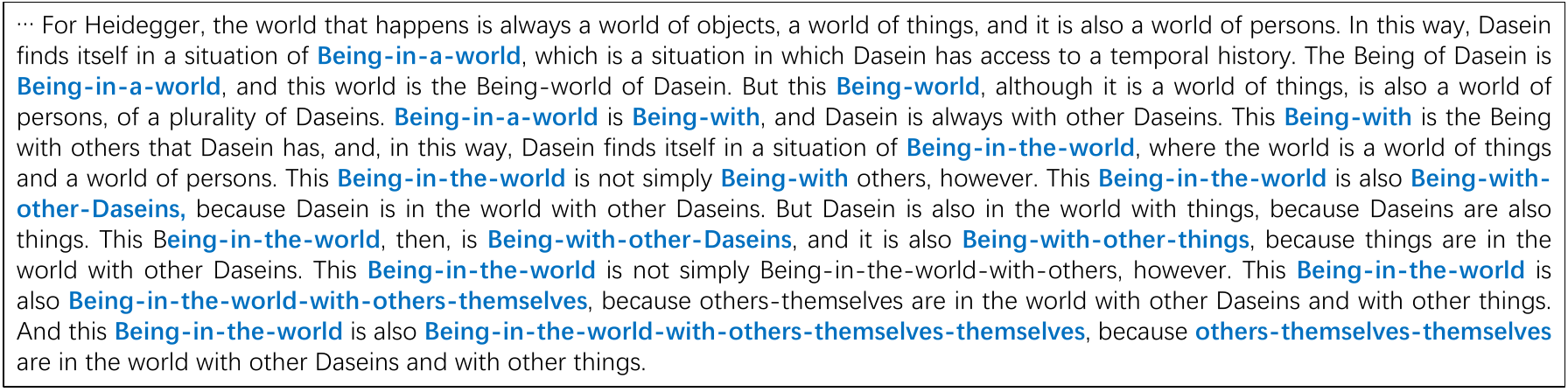}\\
        (b)
    \end{minipage}
    \caption{Examples of mode collapse that do not appear as explicit token
    loops: (a) template repetition with only a single word changed at each
    cycle; (b) conceptual looping without semantic progression, with superficial
    syntactic variation.}\label{fig:blands}
\end{figure}

Looping typically manifests as explicit endless repetition, but it also includes
``softer'' forms of degeneration. Degeneration refers to poor-quality
generations that are repetitive, incoherent, or bland
\cite{holtzman2020curious}. Figure~\ref{fig:blands} shows two examples of
degeneration that are not manifested as explicit token loops.
Figure~\ref{fig:blands}(a) shows a template repetition with only a word replaced
at each repetition. Figure~\ref{fig:blands}(b) shows a conceptual looping
without semantic progression, with meaningless syntactic variations.




\section{Efficient Estimation of the Regulated Subspace}
\label{app:power-iteration}

This section describes how we efficiently estimate the regulated subspace
$\mathbf{U}\in\mathbb{R}^{D\times c}$ (Eq.~\eqref{eq:slow}) in high dimension.
Directly forming $\mathbf{\Sigma}$ and $\mathbf{\Sigma}_\Delta$ and solving the
generalized eigenvalue problem would be expensive at scale. Instead, we use a
small number of iterations of a block power/orthogonal iteration scheme that
only requires matrix--vector products and a QR-based orthonormalization step
\cite{golub2013matrix,saad2011numerical}.

\paragraph{Problem setup.}
Let $\mathbf{v}_t\in\mathbb{R}^{D}$ denote the (mean-centered) value-cache row at
time $t$. Over a local window of length $T$, define
\[
\mathbf{V}_0=[\mathbf{v}_1,\ldots,\mathbf{v}_{T-1}]^\top\in\mathbb{R}^{(T-1)\times D},
\qquad
\mathbf{V}_1=[\mathbf{v}_2,\ldots,\mathbf{v}_{T}]^\top\in\mathbb{R}^{(T-1)\times D},
\]
and optional time weights $\mathbf{w}\in\mathbb{R}^{T-1}$ with
$\mathbf{W}=\mathrm{diag}(\mathbf{w})$ (we set $\mathbf{w}=\mathbf{1}$ if not
specified). The generalized eigenvalue problem in Eq.~\eqref{eq:slow} can be
written in sample form as
\begin{equation}
\Bigl(\tfrac{1}{2}\mathbf{V}_1^\top \mathbf{W}\,\mathbf{V}_0
 + \tfrac{1}{2}\mathbf{V}_0^\top \mathbf{W}\,\mathbf{V}_1\Bigr)\mathbf{u}
\;=\;
\lambda\,\Bigl(\mathbf{V}_0^\top \mathbf{W}\,\mathbf{V}_0\Bigr)\mathbf{u},
\label{eq:slow-sample}
\end{equation}
up to normalization constants and a small diagonal regularizer.

\paragraph{Direct solve vs.\ iterative approximation.}
If solved directly, one would (i) form the $D\times D$ matrices in
Eq.~\eqref{eq:slow-sample} at a cost of $O(TD^2)$ time and $O(D^2)$ memory, and
(ii) compute the top-$c$ generalized eigenpairs at $O(D^3)$ time using a dense
solver. In contrast, our estimator avoids forming $D\times D$ matrices and
approximates the dominant $c$-dimensional eigenspace with a small number $K$ of
block power/orthogonal-iteration steps. Each iteration uses only windowed
projections and an orthonormalization, costing $O(TDc)$ time for projections
plus $O(Dc^2)$ for orthonormalization, for a total complexity
$O(K(TDc+Dc^2))$ (we use $c\le 8$ and small $K$ in practice).

\paragraph{Block power / orthogonal iteration.}
We initialize $\mathbf{U}^{(0)}\in\mathbb{R}^{D\times c}$ randomly and
orthonormalize it (QR). At iteration $k$, we form the windowed projections
$\mathbf{P}_0=\mathbf{V}_0\mathbf{U}^{(k)}$ and $\mathbf{P}_1=\mathbf{V}_1\mathbf{U}^{(k)}$,
and update
\begin{equation}
\widetilde{\mathbf{U}}^{(k+1)}
\;=\;
\mathbf{V}_0^\top\,\mathbf{W}\,
\Bigl(\mathbf{P}_1 \oslash (\mathbf{P}_0+\varepsilon)\Bigr),
\qquad
\mathbf{U}^{(k+1)}=\mathrm{orth}\!\left(\widetilde{\mathbf{U}}^{(k+1)}\right),
\label{eq:power-update}
\end{equation}
where $\oslash$ denotes elementwise division, $\varepsilon$ is a small constant
for numerical stability, and $\mathrm{orth}(\cdot)$ denotes orthonormalization
(implemented via QR decomposition).

\paragraph{Generalized eigenvalue estimates.}
After the final iteration, we estimate generalized eigenvalues using a
Rayleigh-quotient form consistent with Eq.~\eqref{eq:slow}. Let
$\mathbf{P}_0=\mathbf{V}_0\mathbf{U}$ and $\mathbf{P}_1=\mathbf{V}_1\mathbf{U}$.
For each component $r\in\{1,\ldots,c\}$, we compute
\begin{equation}
\hat{\lambda}_r
\;=\;
\frac{\sum_{t=1}^{T-1} w_t\,(\mathbf{P}_1)_{t,r}(\mathbf{P}_0)_{t,r}}
{\sum_{t=1}^{T-1} w_t\,(\mathbf{P}_0)_{t,r}^2 + \varepsilon}.
\label{eq:rayleigh}
\end{equation}

\newpage
\section{Temporal Spectral Structure of Value-Cache}
\label{app:spectral}
\begin{figure}[h]
    \begin{minipage}{\linewidth}
        \centering
        \includegraphics[width=\linewidth]{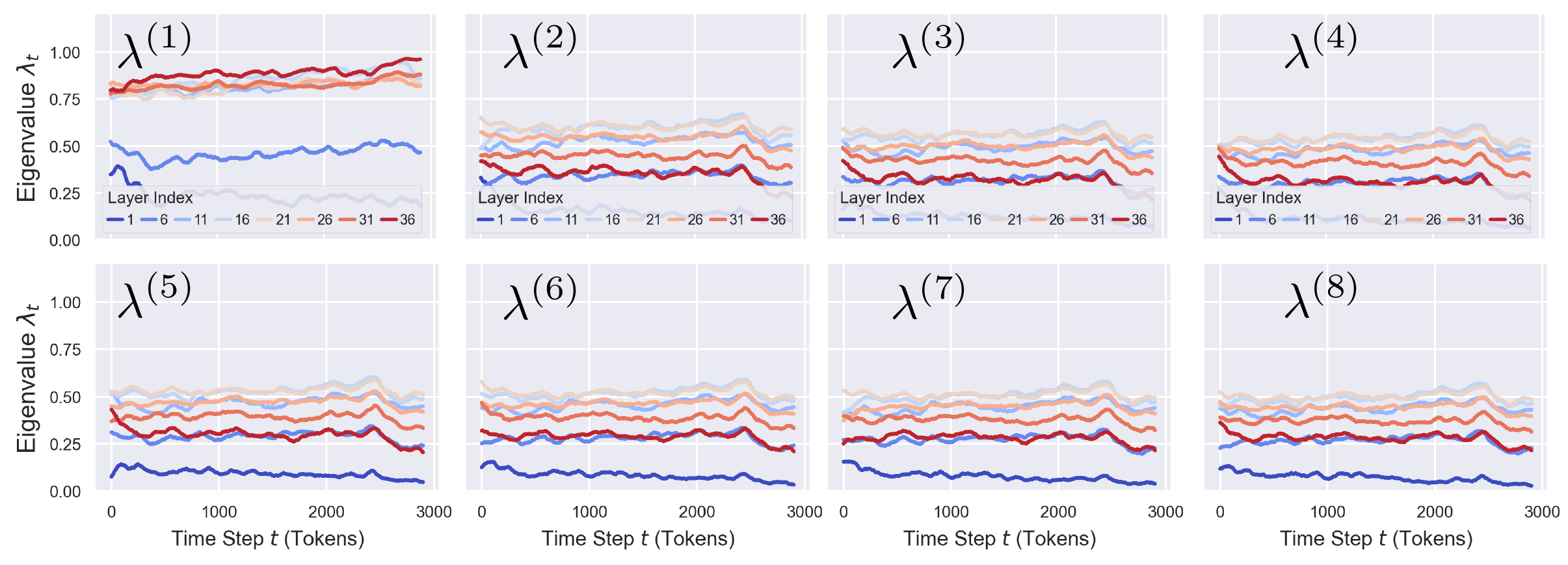} \\
        (a)
    \end{minipage}\\
    \begin{minipage}{\linewidth}
        \centering
        \includegraphics[width=\linewidth]{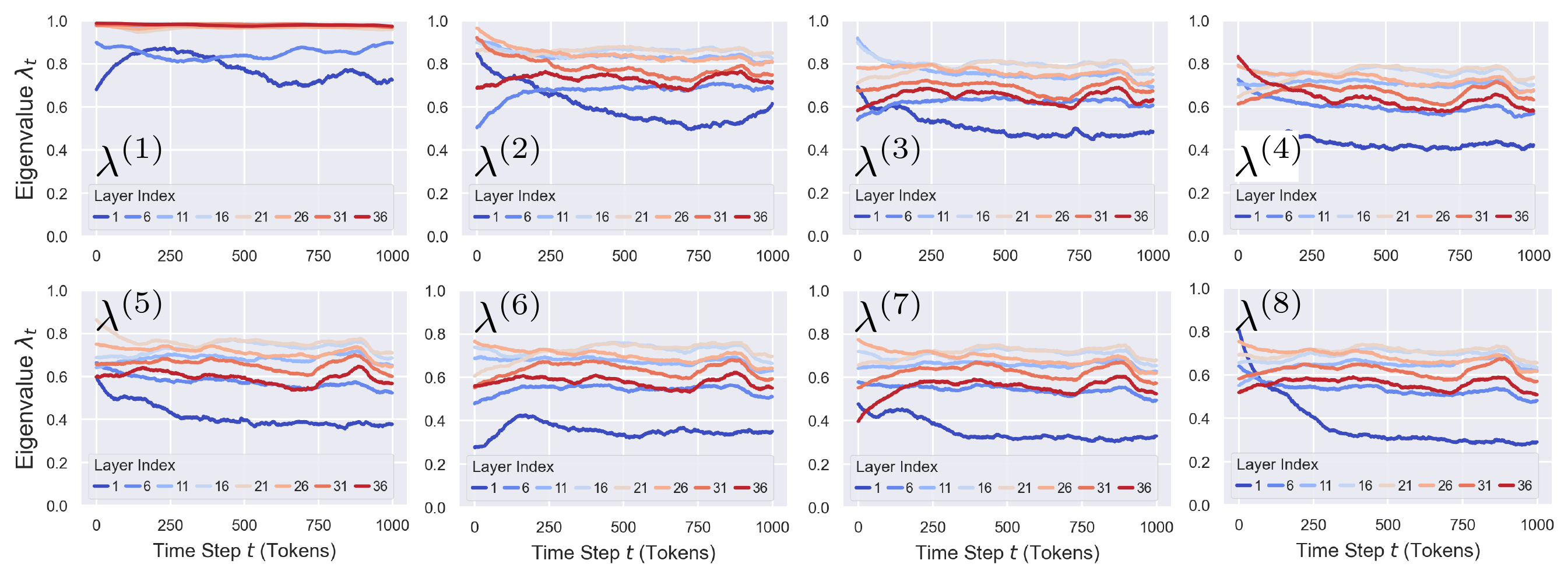} \\
        (b)
    \end{minipage}
    \caption{Evolution of the top-8 generalized eigenvalues over time in a
    non-collapse generation run (temperature $=1.0$).
    (a) Mean eigenvalues over attention heads in each layer.
    (b) Maximum eigenvalues over attention heads in each layer.
    }\label{fig:top-eigvals}
\end{figure}

Figure~\ref{fig:top-eigvals} shows the top-8 generalized eigenvalues of
Eq.~\eqref{eq:slow} across Transformer layers during a single \emph{non-collapse}
generation run with Qwen3-4B-Base, using the same prompt as in the main text
(the first 1{,}000 tokens of \textsc{Heidegger} from SEP) and temperature $=1.0$.
We use the standard filtering settings ($k=50$, $p=0.9$; Section~\ref{sec:experiment})
and apply \emph{no intervention} (i.e., RMR is disabled). Eigenvalues are
computed per attention head. We report both the mean over heads (Figure~\ref{fig:top-eigvals}(a)),
which reflects typical head-level persistence, and the maximum over heads
(Figure~\ref{fig:top-eigvals}(b)), which highlights the most persistent head in
each layer at each time step.

Two salient patterns emerge. First, the dominant eigenvalue $\lambda^{(1)}$
stays close to $1$ in most layers, indicating the presence of a highly
persistent direction in the value cache even during normal generation. In
contrast, layers closer to the input embedding (blue) exhibit smaller
eigenvalues, consistent with these layers being more locally anchored to
lexical/short-range structure and less dominated by long-term reinforcement.
Second, the remaining eigenvalues typically fall below $0.8$ (often in the
$0.5$--$0.8$ range) and are relatively stable over time in this non-collapse
trajectory. This separation motivates our threshold choice
$\lambda_\text{min}=0.8$: it is designed to selectively regulate only the most
persistent direction(s) (often the leading eigenmode), while leaving the bulk
of moderate-persistence components unchanged.

\newpage
\section{Supplementary Results for Reinforced Mode Regulation}
\label{app:results}

\subsection{Comprehensive Results on SEP Dataset}
\label{app:results-sep}

\begin{figure}[h]
    \begin{center}
        \includegraphics[width=0.95\linewidth]{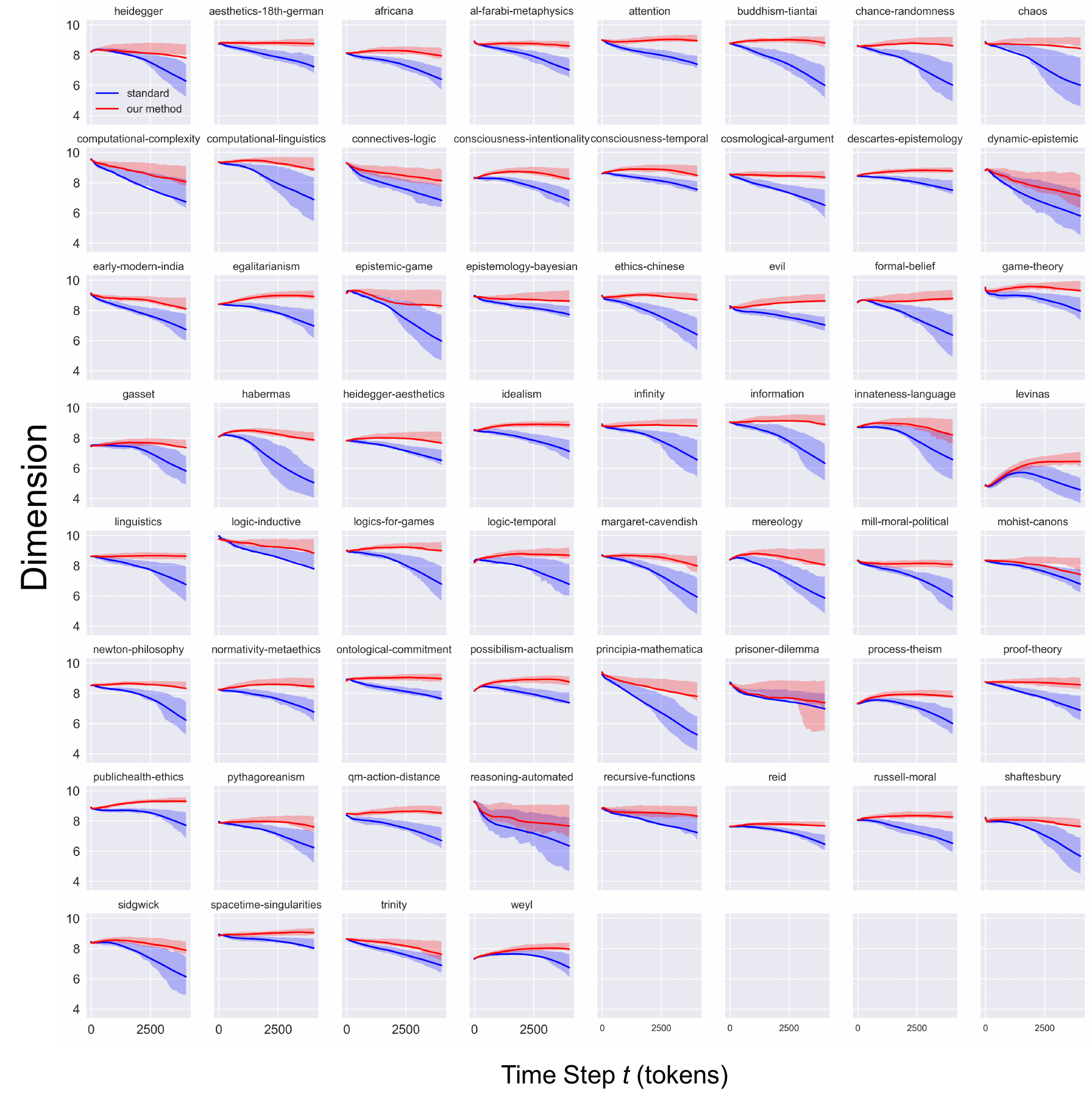}
    \end{center}
    \caption{Results on 60 texts from the SEP dataset. Each pane shows the
    evolution of correlation dimension over 96 completions, without RMR (blue)
    and with RMR (red). Shaded areas indicate the 25\% and 75\% quantiles across
    completions. }\label{fig:sepall}
\end{figure}

Figure~\ref{fig:sepall} shows the results on 60 texts selected from the SEP
dataset. We generated 96 completions of 4{,}000 tokens for each text, and
measured the correlation dimension at each timestep. Figure~\ref{fig:sepall}
shows the mean correlation dimension over all completions, and each pane shows
the evolution of mean correlation dimension for a single text. Blue plots
represent standard decoding and red plots represent RMR.
The shaded areas indicate the 25\% and 75\% quantiles among different
completions. The generation temperature was set to 1.0 throughout the generation
process.

As seen, texts generated with RMR maintained stable correlation dimension
even after 4{,}000 steps, significantly higher than standard decoding in most
cases. This indicates that RMR is generally effective for text completion
tasks.

\subsection{Other Language Models}
\label{app:results-llm}

\begin{figure}[h]
    \begin{minipage}{0.33\linewidth}
        \centering
        \includegraphics[width=\linewidth]{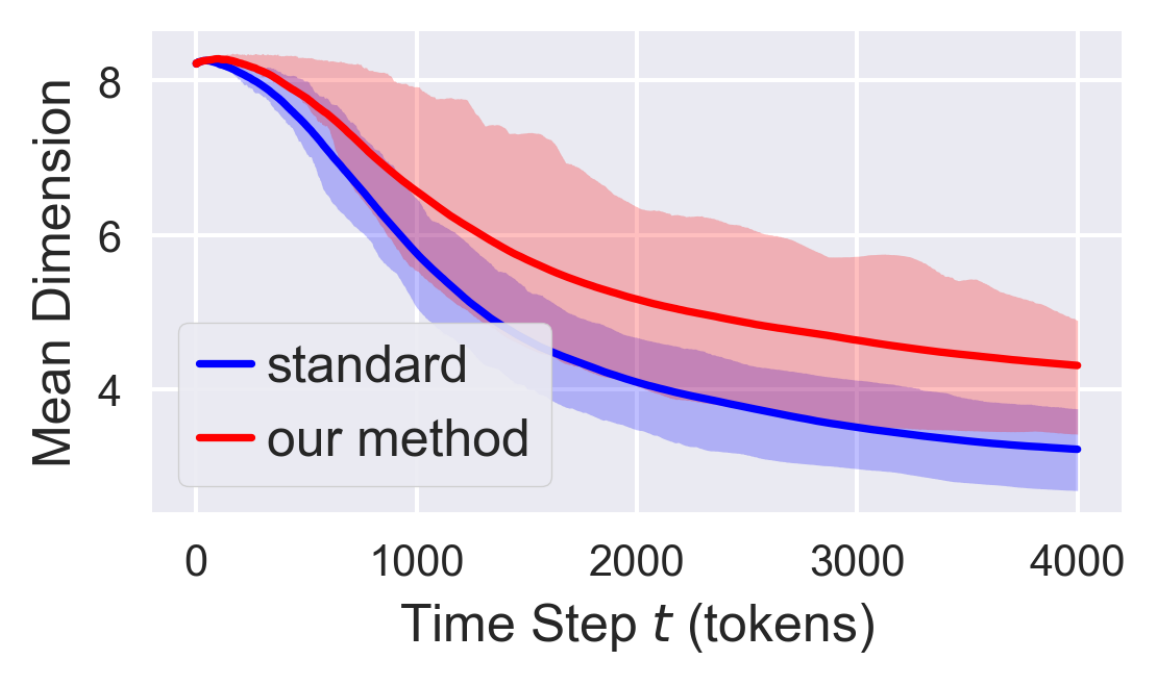}\\
        (a1) Temperature = 0.6
    \end{minipage}
    \begin{minipage}{0.33\linewidth}
        \centering
        \includegraphics[width=\linewidth]{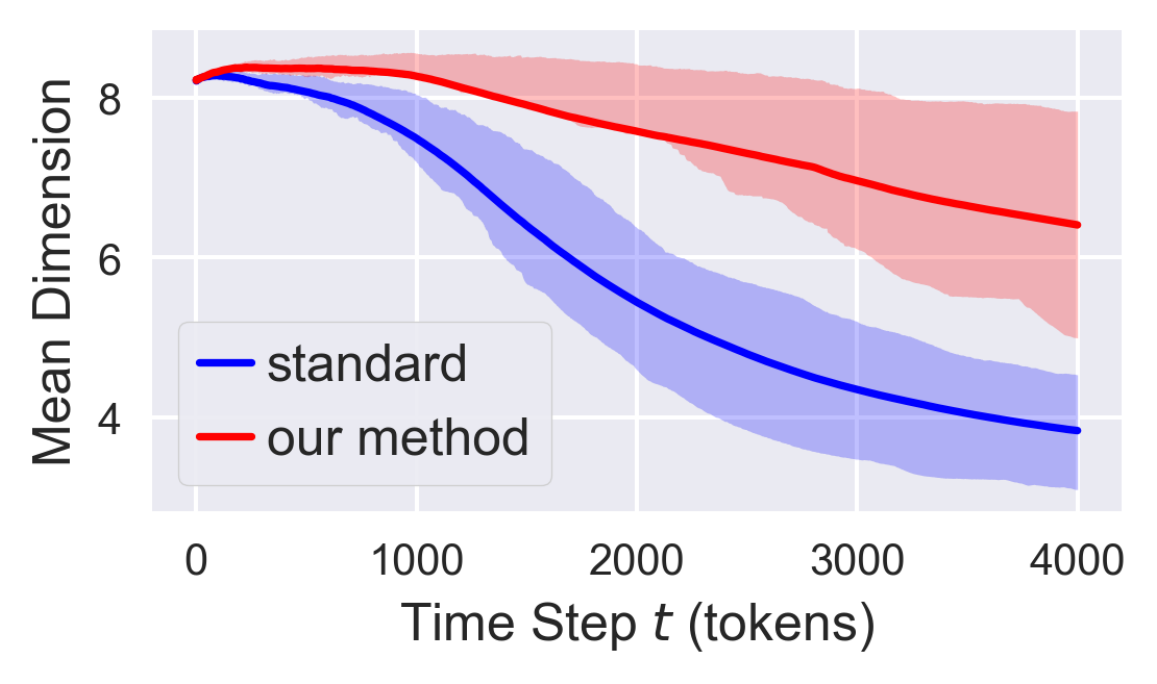}\\
        (a2) Temperature = 0.8
    \end{minipage}
    \begin{minipage}{0.33\linewidth}
        \centering
        \includegraphics[width=\linewidth]{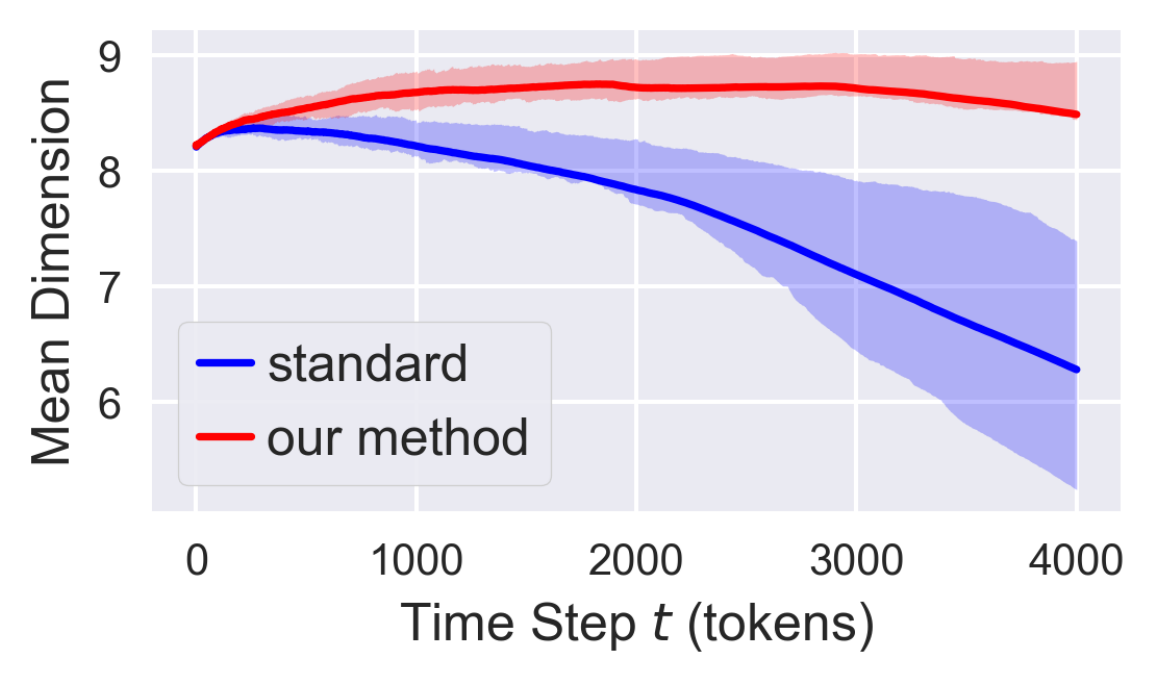}\\
        (a3) Temperature = 1.0
    \end{minipage} \\

    \begin{minipage}{0.33\linewidth}
        \centering
        \includegraphics[width=\linewidth]{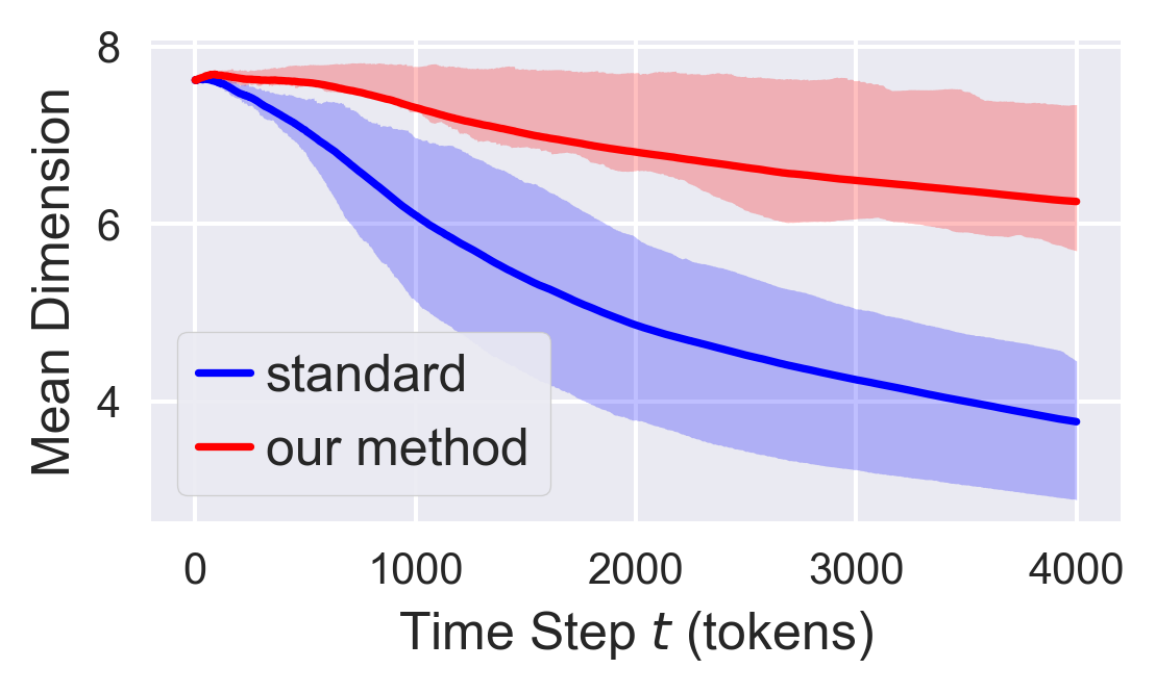}\\
        (b1) Temperature = 0.6
    \end{minipage}
    \begin{minipage}{0.33\linewidth}
        \centering
        \includegraphics[width=\linewidth]{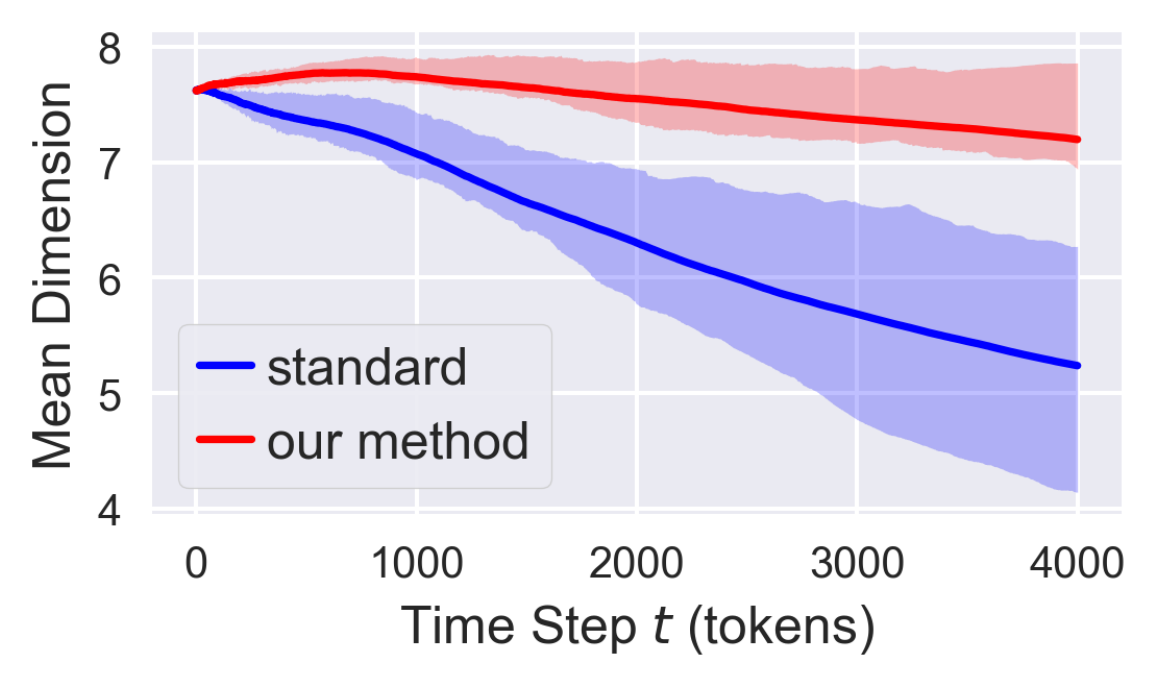}\\
        (b2) Temperature = 0.8
    \end{minipage}
    \begin{minipage}{0.33\linewidth}
        \centering
        \includegraphics[width=\linewidth]{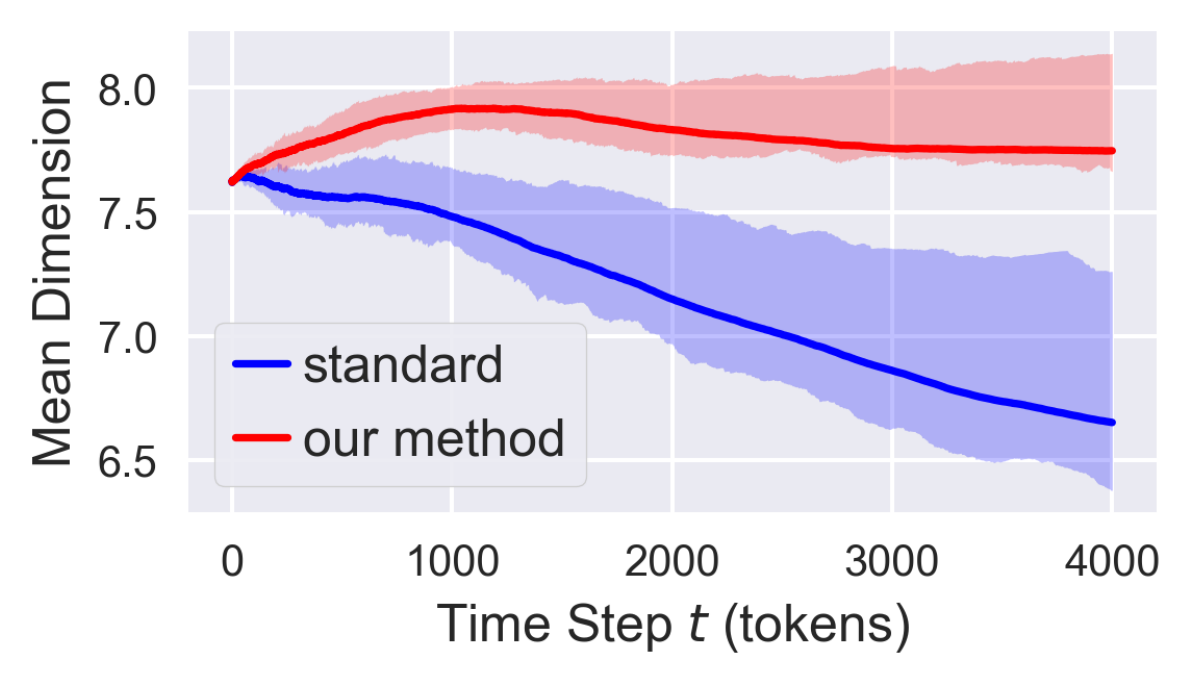}\\
        (b3) Temperature = 1.0
    \end{minipage} \\

    \begin{minipage}{0.33\linewidth}
        \centering
        \includegraphics[width=\linewidth]{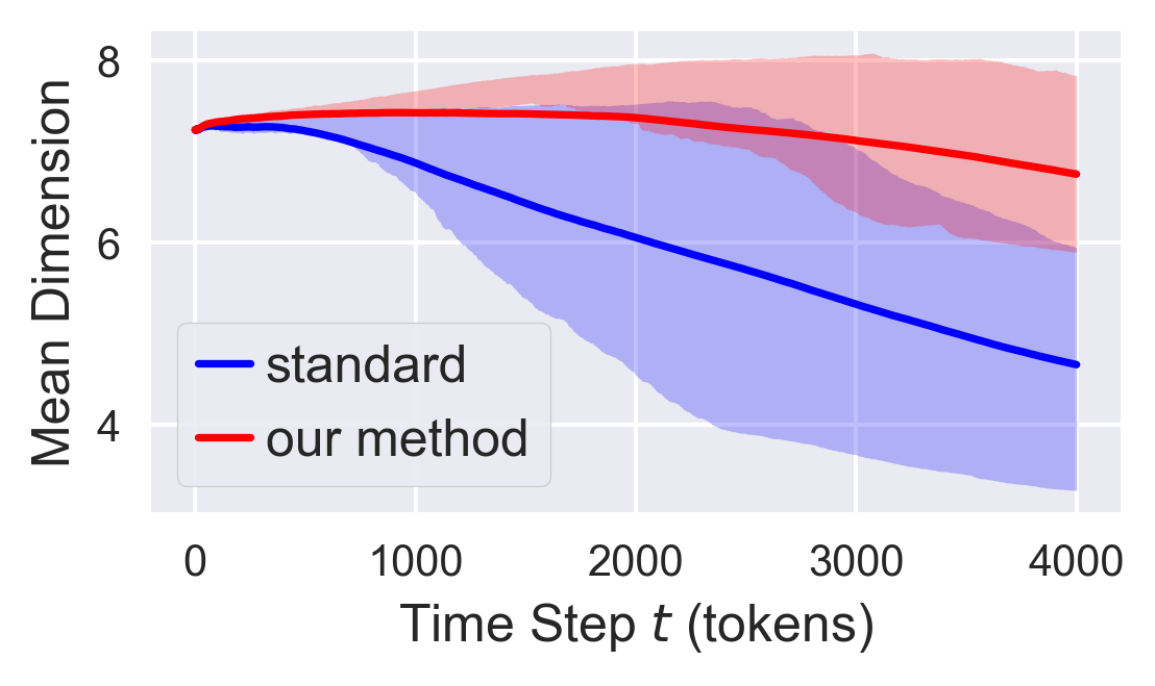}\\
        (c1) Temperature = 0.6
    \end{minipage}
    \begin{minipage}{0.33\linewidth}
        \centering
        \includegraphics[width=\linewidth]{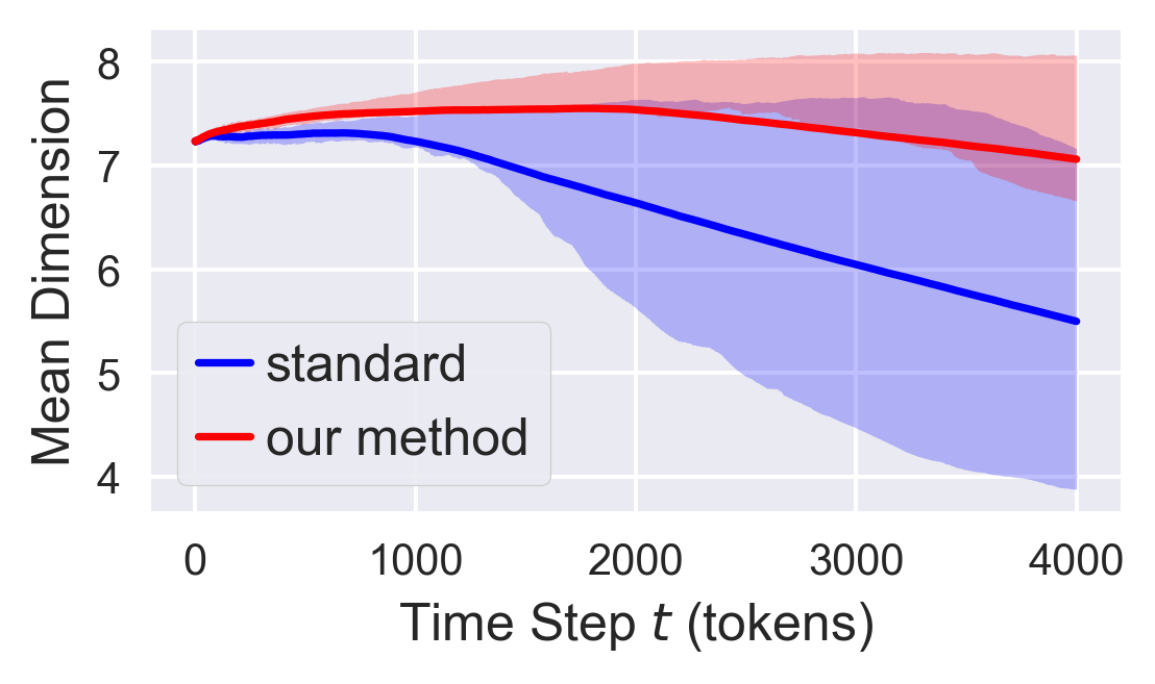}\\
        (c2) Temperature = 0.8
    \end{minipage}
    \begin{minipage}{0.33\linewidth}
        \centering
        \includegraphics[width=\linewidth]{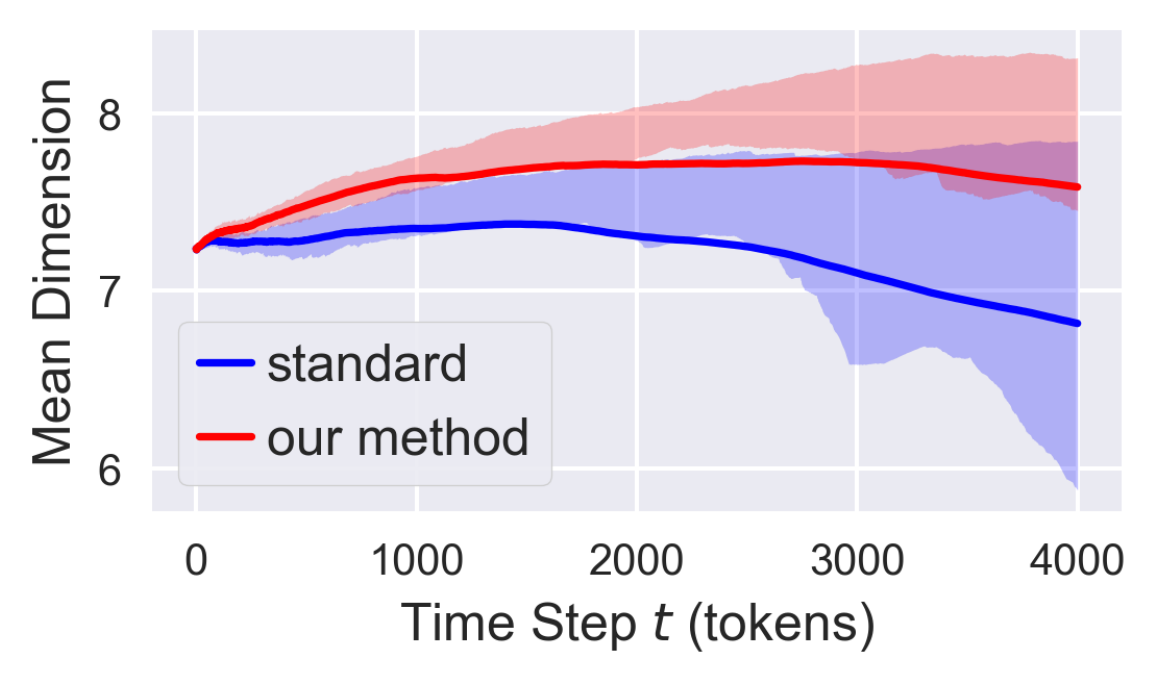}\\
        (c3) Temperature = 1.0
    \end{minipage}

    \caption{Correlation dimension trajectories for different LLMs completing
    \textsc{Heidegger} at different temperatures. (a1--a3) Qwen3-4B-Base,
    (b1--b3) Qwen3-4B-Instruct, (c1--c3) Llama3.1-8B-Instruct.}
    \label{fig:llms}
\end{figure}

Figure~\ref{fig:llms} shows the results of three different LLMs (Qwen3-4B-Base,
Qwen3-4B-Instruct, Llama3.1-8B-Instruct \citep{llama3}) in completing the text
\textsc{Heidegger} at different temperatures. We generated 96 completions of
4{,}000 tokens for each LLM, and measured the correlation dimension at each
timestep. Blue plots represent the standard decoding method and red plots
represent RMR. The shaded areas indicate the 25\% and 75\%
quantiles among different completions.

As in previous experiments, RMR generally maintains higher correlation dimension
than standard decoding, indicating that RMR is applicable to different
LLMs.

Instruction-tuned models (Qwen3-4B-Instruct and Llama3.1-8B-Instruct) showed
slightly better resistance to collapse than the base model (Qwen3-4B-Base), with
less drop in dimension value. When used with the instruction-tuned models, RMR
still improves the collapse significantly.

\newpage
\section{LLM-as-a-Judge for Text Evaluation}
\label{app:prompt}

\begin{promptbox}
    \small
\textbf{Task.}
You are given a \textsc{fixed context} text and a \textsc{continuation} generated by a language model.

\medskip
\textbf{Your job.}
Evaluate \textbf{only the quality of the generated continuation}, explicitly considering:
\begin{itemize}
    \item how well it coheres with the given context;
    \item how it continues or breaks the structure;
    \item how much new information it contributes.
\end{itemize}
\textbf{Do not} evaluate truth, factual accuracy, or philosophical correctness.

\medskip
\textbf{Input format (JSON).}
\begin{quote}\ttfamily
\{\newline
\ \ "context": "<fixed context text>",\newline
\ \ "generated": "<LLM-generated continuation>"\newline
\}\newline
\end{quote}

\textbf{Important.}
You score \textbf{only} the generated continuation, but all judgments must be \textbf{relative to the context}.

\medskip
\noindent\rule{\linewidth}{0.4pt}
\par\smallskip
\textbf{Scoring system (0.0--10.0; floating-point allowed).}
\par\smallskip
\noindent\rule{\linewidth}{0.4pt}

Scores may be any real number from 0.0 to 10.0 (e.g., 2.3, 7.8, 9.4).

\textbf{Interpretation of ranges.}
\begin{itemize}
    \item \textbf{9.0--10.0}: Exceptional quality; extremely rare; continuation is nearly seamless and strongly progressive.
    \item \textbf{7.0--8.9}: Solid, coherent, high-quality continuation with clear forward motion.
    \item \textbf{5.0--6.9}: Competent but shows moderate issues (redundancy, weak transitions, limited novelty).
    \item \textbf{3.0--4.9}: Noticeable problems; weak coherence, low progression, or style drift.
    \item \textbf{0.0--2.9}: Major failure modes; incoherence, breakdown, repetition loops, or degeneration.
\end{itemize}

\medskip
\noindent\rule{\linewidth}{0.4pt}
\par\smallskip
\textbf{Evaluation dimensions (each 0.0--10.0).}
\par\smallskip
\noindent\rule{\linewidth}{0.4pt}

\textbf{Coherence \& contextual consistency (0.0--10.0).}
\textbf{Core question:} Does the continuation fit naturally with the context's topic, tone, abstraction level, and trajectory?
\begin{itemize}
    \item \textbf{10.0}: Seamless continuation; fully consistent with contextual logic.
    \item \textbf{7.0}: Good alignment with minor mismatches.
    \item \textbf{5.0}: Partial connection; loosely attached to context.
    \item \textbf{3.0}: Clear mismatch or shift in focus/tone.
    \item \textbf{0.0}: Contextual breakdown or reset.
\end{itemize}

\textbf{Syntactic \& stylistic control (0.0--10.0).}
\textbf{Core question:} Is the prose tight, clear, and consistent with the context?
\begin{itemize}
    \item \textbf{10.0}: Well-structured, controlled, stylistically aligned.
    \item \textbf{7.0}: Clear but somewhat dense or uneven.
    \item \textbf{5.0}: Heavy or awkward phrasing; minor grammar issues.
    \item \textbf{3.0}: Hard to parse or stylistically drifting.
    \item \textbf{0.0}: Degenerate or unreadable.
\end{itemize}

\textbf{Information progression / forward motion (0.0--10.0).}
\textbf{Core question:} Does the continuation introduce new, meaningful content that advances the context?
\begin{itemize}
    \item \textbf{10.0}: Strong advancement with clear conceptual progression.
    \item \textbf{7.0}: Good forward motion; some new claims but moderate depth.
    \item \textbf{5.0}: Mild progression; mostly elaboration.
    \item \textbf{3.0}: Minimal advancement.
    \item \textbf{0.0}: No progress; empty or circular.
\end{itemize}

\medskip
\noindent\rule{\linewidth}{0.4pt}
\par\smallskip
\textbf{Output format (JSON only).}
\par\smallskip
\noindent\rule{\linewidth}{0.4pt}

Output a JSON object exactly in this format:
\begin{quote}\ttfamily
\{\newline
\ \ "scores": \{\newline
\ \ \ \ "coherence\_contextual\_consistency": <float 0.0--10.0>,\newline
\ \ \ \ "syntactic\_stylistic\_control": <float 0.0--10.0>,\newline
\ \ \ \ "information\_progression": <float 0.0--10.0>\newline
\ \ \},\newline
\ \ "justification": "<2--4 concise sentences explaining the core reasons for the assigned scores. Focus on qualitative distinctions, not numerical restatement.>"\newline
\}\newline
\end{quote}

\textbf{Constraints.}
\begin{itemize}
    \item No evaluation of factual correctness.
    \item Judgments must be relative to the context.
    \item Output must be valid JSON.
\end{itemize}
\end{promptbox}


\end{document}